%% file: main.tex
\documentclass{article}
\usepackage{subfigure}
\usepackage{amsmath}

\usepackage{arxiv}

\usepackage[utf8]{inputenc} 
\usepackage[T1]{fontenc}    
\usepackage{hyperref}       
\usepackage{url}            
\usepackage{booktabs}       
\usepackage{amsfonts}       
\usepackage{nicefrac}       
\usepackage{microtype}      
\usepackage{lipsum}
\usepackage{graphicx}
\graphicspath{ {./images/} }

\title{VIPPrint: A Large Scale Dataset of Printed and Scanned Images for Synthetic Face Images Detection and Source Linking}

\author{
 Anselmo Ferreira\\
  Department of Information Engineering \\ and Mathematics
  \\ University of Siena 
  \\ Siena SI, Italy 53100 \\
  \texttt{anselmo.castelo@unisi.it} \\
   \And
  Ehsan Nowroozi  \\
  Department of Information Engineering \\ and Mathematics
  \\ University of Siena 
  \\ Siena SI, Italy 53100 \\
  \texttt{ehsan.nowroozi65@gmail.com} \\
  \And
 Mauro Barni \\
  Department of Information Engineering \\ and Mathematics
  \\ University of Siena 
  \\ Siena SI, Italy 53100 \\
  \texttt{barni@diism.unisi.it} \\
}

\begin{document}
\maketitle
\begin{abstract}
The possibility of carrying out a meaningful forensics analysis on printed and scanned images plays a major role in many applications. First of all, printed documents are often associated with criminal activities, such as terrorist plans, child pornography pictures, and even fake packages. Additionally, printing and scanning can be used to hide the traces of image manipulation or the synthetic nature of images, since the artifacts commonly found in manipulated and synthetic images are gone after the images are printed and scanned. A problem hindering research in this area is the lack of large scale reference datasets to be used for algorithm development and benchmarking. Motivated by this issue, we present a new dataset composed of a large number of synthetic and natural printed face images. To highlight the difficulties associated with the analysis of the images of the dataset, we carried out an extensive set of experiments comparing several printer attribution methods. We also verified that state-of-the-art methods to distinguish natural and synthetic face images fail when applied to print and scanned images. We envision that the availability of the new dataset and the preliminary experiments we carried out will motivate and facilitate further research in this area.
\end{abstract}

\keywords{Digital Image Forensics; Printer Source Attribution; Printed Image Tampering Detection.}

\input{introduction}

\input{related-work}
\input{dataset}

\input{experimental-setup}

\input{experiments}

\input{conclusion}

\bibliographystyle{unsrt}  
\bibliography{references}  

\end{document}

%% file: introduction.tex
\section{Introduction}

The abundant availability of new technologies for generating physical documents such as printers and scanners has raised many concerns about their misuse, being generating illegal documents,  misguiding investigations through the generation of fake evidence, or even hiding relevant traces in a criminal investigation. For instance, child pornography pictures can be printed and distributed between pedophiles in order to avoid virtual monitoring from the police, and illegal amendments can be incorporated in printed contracts without previous notice. Furthermore, professional printers can be used to print fake currency and packages of fake products, causing several negative effects on the economy. Finally, printing and scanning can be used to hide the traces of image manipulation or the synthetic nature of the images, since the artifacts commonly found in manipulated and, synthetic images are no more present or detectable after the images have been printed and scanned.

As a countermeasure to the diffusion of counterfeited printed documents, most of the major manufacturers of color laser printers have signed a secret agreement with governments to let the printers include secret (invisible) yellow dots into the printed documents \cite{eff:nodots}. Such dots, also called Machine Identification Code (MIC) or simply printer steganography, are used to identify the source of printed documents, as unique yellow-dots patterns are used to identify different source printers. However, such a feature is not enabled in all laser printers and, as shown in \cite{richteretal:2018}, the yellow-dots patterns can be easily anonymized, leaving the authentication of printed documents problem unsolved.

The challenges posed by printed documents forensics have pushed the multimedia forensics research community to look for viable solutions based on the analysis of the artifacts left by the printers into the printed documents. In general, printed document forensics can be split into three main research areas: (i) source linking (a.k.a. printer attribution); (ii) detection of printed manipulated images; and (iii) detection of printed and scanned synthetic images. Solutions for printer attribution are mostly based on the analysis of the extrinsic artifacts contained in printed documents, with the most popular ones for laser printers being the banding, jitter, and skewed jitters. The presence of these artefacts has been exploited by several works to identify the source of printed texts \cite{alietal:2004,mikkilinenietal:2004,Mikkilinenietal:2005,Mikkilinenietal:2005a,keeandfarid:2008,dengetal:2008,wuetal:2009,jiangetal:2010,mikkilinenietal:2011,tsaiandliu:2013,elkasrawishafait:2014,tsaietal:2014,haoetal:2015,shangetal:2015,tsaietal:2015,anselmoetal:2017,joshikhanna:2018,joshietal:2018,Jainandkhanna:2019,joshikhanna:2020}, color images \cite{alietal:2003,eidetal:2008,bulanetal:2009,choietal:2009,choietal:2010,ryuetal:2010,tsaielat:2011,choietal:2013,kimandlee:2014,wuetal:2015,kimandlee:2015} or both \cite{anselmoetal:2015,tsaietal:2017,bibietal:2019}. Manipulation detection in printed documents has received some attention only recently \cite{jamesetal:2020} and mainly refers to unveiling post-processing operations that could alter the semantic meaning of the images, and usually exploit texture descriptors and deep neural networks to identify the visual artifacts introduced by such manipulations. Finally, as far as we know, despite the intense research devoted to the detection of images generated by Generative Adversarial Networks (GANs) \cite{marraetal:2018,lakshmananetal:2019,barnietal:2020}, scarce attention has been paid to the detection of such images in printed documents.

Notwithstanding the research carried out so far, progresses in this area are hindered by the lack of large reference datasets. The few existing datasets, in fact,  present at least one of the following issues: (i) they contain \textit{ad-hoc} data prepared for specific research only; (ii) the printed patterns are often simple ones, like icons, texts, and halftone patterns; (iii) most of them consider old and non-professional printers; (iv) they do not consider copies of the same printer brand and model; and (v) to the best of our knowledge, no dataset with printed complex fake images exist. This last issue is particularly important and challenging, as most of the artifacts used to detect image manipulations, such as the correlation between RGB channels, discrete cosine transform irregularities, and even illumination inconsistencies are gone after the images are printed and scanned. This problem is worsened by the observation that printing and scanning back a manipulated image is one of the most powerful and simplest attacks an adversary can conceive to fool manipulation detectors. The availability of a large reference dataset overcoming the above problems may be of great help to foster new advances in printed documents investigations, concerning both the detection of manipulated and synthetic documents and the attribution of printed documents to the device that generated them.  

In this paper, we aim at filling these gaps by presenting a large scale dataset that can be used for both applications: source attribution and synthetic images detection. Due to their relevance in image forensics applications, the dataset focuses on face images. In particular, the initial version of the dataset (we are planning to update it continuously in the next years) is composed of images printed by several printers and scanned back with a high-quality scanner. The images in the dataset are divided into (i) pristine images, to be used for the source attribution problem; and (ii) synthetic face images generated by three different Generative Adversarial Networks (GANs). The dataset is further split into several subsets, containing regions of interest with different sizes for the investigation of localized artifacts. To evaluate the difficulties associated with the forensics analysis of the images contained in the dataset, we carried out an extensive comparative study including several source attribution and synthetic image detection baseline methods.

In summary, the contributions of this paper are:

\begin{enumerate}
    \item We present a large scale dataset of color printed face images for digital image forensics purposes, such as source attribution and synthetic images detection (deep fake images).   
    \item We increase the diversity of the images in our dataset to make it suitable for approaches working on images of different sizes. To do that, we make available full scanned images and regions of interest with different sizes.
    \item To the best of our knowledge, our dataset is the first large scale dataset with printed and scanned artificial images created with GANs such as StyleGAN2 \cite{karrasetal:2019}, ProgressiveGAN \cite{karrasetal:2017}, and StarGAN \cite{choietal:2017}.
    \item We present the results of an in-depth comparative study conducted on the new dataset regarding several baseline approaches, including both data-driven methods and methods based on handcrafted features. The comparison regards both source attribution and synthetic image detection.   
\end{enumerate}

The rest of this paper is organized as follows: in Section \ref{related-work} we report some related work and discuss the limitations of datasets used in the literature. In Section \ref{dataset}, we present our dataset and several configurations considered to generate data. In Section \ref{experimental-setup}, we discuss the experimental setup considered to assess the difficulty of such a dataset. Finally, Section \ref{experiments} reports the achieved results and, in Section \ref{conclusion}, we conclude this paper and discuss the future work that we are aiming to do in such a dataset.

%% file: related-work.tex
\section{Related work}
\label{related-work}



Several works have investigated the exploitation of the artifacts left by the printers into the printed documents to identify their source. Here we focus our works aiming at source linking after document scanning, as they are usually cheaper, non-destructive, and fast. 

Common surveys in the literature \cite{khannaetal:2006,khannaetal:2008,chiangetal:2009,devietal:2010} divide source linking methods according to the kind of documents they focus on, namely: printed text documents,  printed color image documents, or both. Moreover, we can distinguish between methods aiming at identifying the technology used to print the documents, \textit{e.g.,} inkjet, laser, \textit{etc.}, and those trying to link the printed document to the single device that was used to print them \cite{oliverchen:2002,lampertetal:2006,schulzeetal:2009,schreyeretal:2009,ankushetal:2010}. In this section, we briefly review the second class of methods since research in that area is more advanced.

Generally speaking, there are two kinds of clues in printed documents that could guide a forensic investigation aimed at identifying the specific source of the document: \textit{intrinsic} and \textit{extrinsic} signatures. Intrinsic signatures are introduced by the printing process itself, whereas extrinsic signatures are intentionally inserted into the printed material. Three of the most investigated intrinsic signatures in laser printers are banding, jitter, and skewed jitters. Eid \textit{et al.} \cite{eidetal:2008} characterized banding as a textural pattern composed of horizontal, low frequency, and periodic artifacts caused by the laser printer components variation, vibration, and speed regulation that can uniquely identify different printers. Similarly, the jitter consists of horizontal artifacts, but with a different frequency range and duration, and is caused by oscillatory disturbances of the printer's drum and the developer roller. Finally, skewed jitter is also a periodic artifact like the others, but it differs from the previous ones as it is formed by vertical lines. With regard to extrinsic signatures, some relevant works include embedding code sequences in electrophotographic halftone images \cite{chiangetal:2011} and also Machine Identification Codes \cite{Beusekometal:2012}. Approaches based on extrinsic signatures require expensive modifications in the printing device, and also some of these extrinsic signatures can even be erased from the printed material \cite{richteretal:2018}. 

With regard to the attribution of printed texts (\textit{i.e.,} black and white dots only), most of the techniques based on intrinsic signatures treat such a problem as a \textit{texture identification problem}, as artifacts such as banding are not easy to be obtained from text \cite{choietal:2009}. For this set of techniques, the same patterns are extracted from the documents and subtle differences among them can be discriminated when printed by different printers \cite{anselmoetal:2015}. One of the pioneers works in this regard comes from Ali \textit{et al.} in 2004 \cite{alietal:2004}. The authors consider the pixel values of letters "I" as features in a multi-class classification problem. After the letters are classified, the source of a document can be found by verifying the most voted class among all the individual letters "I" classification. Several other techniques used a similar pipeline with few modifications, such as considering the statistics of gray-level co-occurrence matrices \cite{mikkilinenietal:2004, Mikkilinenietal:2005,Mikkilinenietal:2005a, mikkilinenietal:2011}, Distance Transform \cite{dengetal:2008}, Discrete Cosine Transform \cite{jiangetal:2010}, statistics of gray-level co-occurrence matrices together with residual noise and sub-bands of Wavelet Transform  \cite{tsaiandliu:2013,tsaietal:2014,elkasrawishafait:2014,tsaietal:2015}, deep neural networks \cite{anselmoetal:2017,joshietal:2018}, \textit{ad-hoc} texture descriptors \cite{joshikhanna:2018,joshikhanna:2020} among others \cite{keeandfarid:2008,wuetal:2009,haoetal:2015,shangetal:2015,Jainandkhanna:2019}. 

A second set of techniques to identify the source of any printed document focuses on the intrinsic signatures of documents containing colored pictures. In this case, banding artifacts are more evident as more patterns are printed (including the background). In this regard, one of the pioneer works is the one from Ali \textit{et al.} \cite{alietal:2003}, where the authors proposed to capture banding artifacts by applying the Fourier transform in image patches to get different banding frequencies. Eid \textit{et al.} \cite{eidetal:2008} applied a similar strategy to jitter artifacts by using Gabor filtering and Discrete Fourier Transform.  

Another set of works on colored documents source linking treat intrinsic artifacts as noise. Choi \textit{et al.} \cite{choietal:2009} discriminate printers by calculating 39 noise features from the diagonal (HH) sub-band of the discrete wavelet transform in pairwise and individual RGB and CMYK channels. In a subsequent work by the same authors \cite{choietal:2010}, noise is estimated after Wiener filtering and Gray Level Co-occurrence Matrix statistics. Tsai \textit{et al.} \cite{tsaielat:2011} calculate 45 statistics in HH, LH and HL sub-bands of Discrete
Wavelet Transform with further feature selection. Choi \textit{et al.} \cite{choietal:2013} extend their previous work in \cite{choietal:2009} by estimating noise with Wiener filtering and 2D Discrete Wavelet Transform, characterizing them with 384 statistical filters on gray level co-occurrence matrices that describe single channels of residual images and pair-wise channels. Other important techniques for color documents source attribution involve describing geometric distortions \cite{bulanetal:2009,wuetal:2015} and halftone texture descriptors \cite{ryuetal:2010,kimandlee:2014,kimandlee:2015}.

Finally, a bunch of techniques aims at identifying the source of printed documents regardless of their content. Ferreira \textit{et al.} \cite{anselmoetal:2015} propose an extension of the Gray Level Co-occurrence Matrices descriptor considering
more directions and scale and also a new descriptor, called Convolutional Texture Gradient Filter, that builds histograms of filtered textures with specific gradients intervals. The authors validated these approaches not only on letters "E" of printed texts but also on regions of interest called frames, which are rectangular areas with sufficient printed material, being them images, texts, or both. Bibi \textit{et al.} \cite{bibietal:2019} use a similar strategy using chunks of printed materials, but their solution involves convolutional neural networks. Finally, Tsai \textit{et al.} \cite{tsaietal:2017} apply nine different filters, fusing several previous strategies such as extracting features from gray-level co-occurrence matrices, Discrete Wavelet Transform, spatial filters, Gabor filter, Wiener filter, Gray Level Co-occurrence Matrices features, and fractal features. 

Notwithstanding the abundant development of source linking approaches for printed documents, the identification of image forgeries and synthetic images from a printed and scanned version of a digital image has received considerably less attention. One of the few works in this area has been published in \cite{jamesetal:2020}, where simple print and scan attacks of manipulated printed documents with recompression, filtering, noise addition, and other simple image operations are detected by a specialized CNN architecture.

Therefore, although printed document forensics (especially printer source attribution) has received much attention in the last years, there are still several issues to be tackled before solutions applicable to real-world scenarios are developed. Among them, the following two issues are relevant for the present paper:

\begin{enumerate}
    \item There is a need for a publicly available dataset that grows through time to include modern printers with different technologies and manufacturing procedures. We expect that different printers manufactured at different times generate different artifacts in printed documents that cannot be detected by previous works.
    \item There is a need for multimedia forensic techniques able to detect deepfake printed images. This is a very challenging problem since several artifacts such as the correlation between RGB channels, discrete cosine transform irregularities, and even illumination inconsistencies in the digital image versions are usually removed by the print and scan process. In this way, although several adversarial attacks have been discussed in the literature \cite{NOWROOZI2021102092}, the print and scan procedure is the easiest yet most powerful attack an adversary could perform against deepfake digital image detectors.  
\end{enumerate}

Therefore, the present work aims at moving a first step towards the solution of the above problems. This is done by presenting a long term dataset addressing both tasks: real-world source attribution with modern printers, and deepfake detection in printed and scanned documents. The details of the dataset we have constructed are described in the following sections.

%% file: dataset.tex
\section{The VIPPrint dataset}\
\label{dataset}

In this work, we present a new dataset trying to minimize some of the issues of existing datasets. The new dataset, which we call \textit{VIPPrint}\footnote{The dataset is named after the VIPP group}, consists of two sections. The first one focuses on printer source attribution and solves some common limitations in previous works, such as (i) \textit{lack of diversity} and (ii) lack of redundancy. Concerning the lack of diversity, the dataset contains printers of different models and printing resolutions. This is an important issue when considering source attribution in real-world applications such as anti-counterfeiting detection, where the printing resolution used for printing a counterfeited document or package is unknown. The inclusion in the dataset of diverse printers marks a significant difference concerning existing datasets, which usually look at artifacts associated with specific printing technologies at fixed resolutions. As to lack of redundancy, very few works have analyzed the effect of the presence of two or more printers of the same model and brand in the dataset, thus neglecting the overlapping effect associated with the presence of two identical printers. The second section of the dataset considers an important, yet understudied, problem in digital image forensics: the detection of synthetic fake images such as those created by Generative Adversarial Networks after print and scan.

The importance of the new dataset for digital image forensics is twofold: (i) it can foster the development of novel solutions for digital image forensics capable to withstand a print and scan procedure, and (ii) it can inspire new techniques for source attribution of fake colored documents printed by modern printers, thus linking the fake content to the owner, or user, of the printer.

Concerning the content of the images composing the dataset, we decided to consider face images. The first reason for such a choice is that face images are particularly relevant in many applications related to biometric recognition, criminal investigations, and misinformation. A second reason is the availability of large scale datasets of face images that can be used as a starting point for the construction of the printed and scanned dataset. Giving researchers the possibility to work both with the digital images and their printed and scanned versions can represent an added value in many applications. Finally, AI-based techniques to generate synthetic images are particularly advanced in the case of face images, whose quality has reached unprecedented levels with no or very few semantic artifacts the forensic analysis can rely on \cite{karrasetal:2019}.

The details of the two sections the VIPPrint dataset consists of are discussed in the following. 

\subsection{VIPPrint Dataset for Source Attribution}

\begin{figure}[!ht]
    \centering
    \subfigure{\includegraphics[scale=0.08]{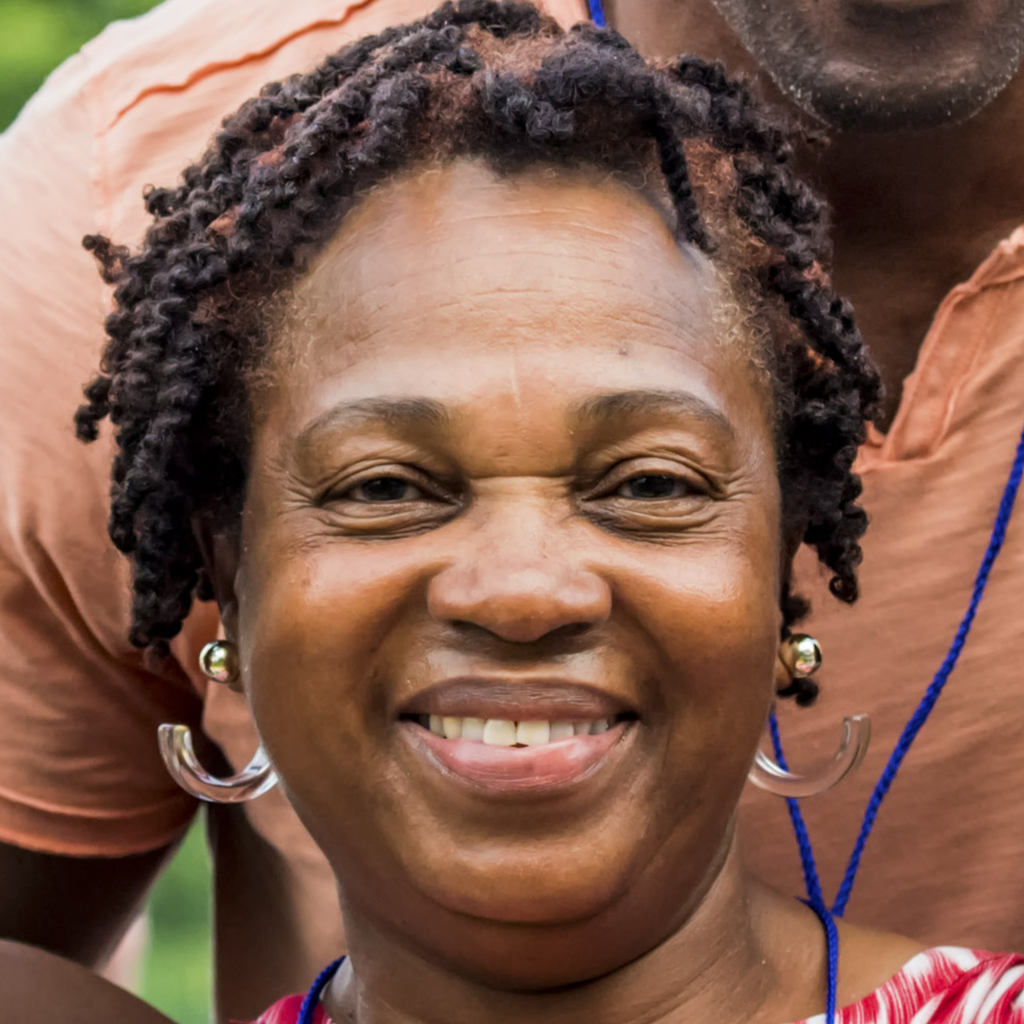}}
    \subfigure{\includegraphics[scale=0.08]{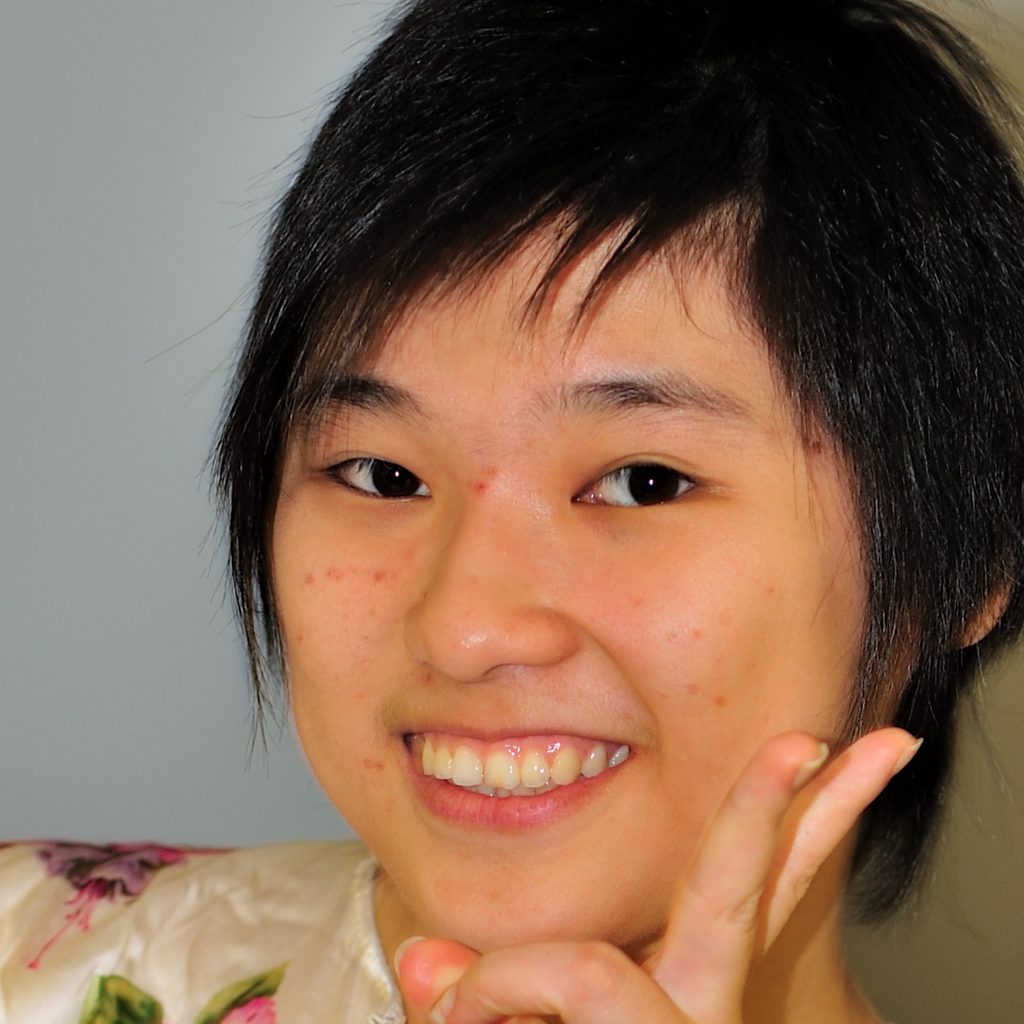}}
    \subfigure{\includegraphics[scale=0.08]{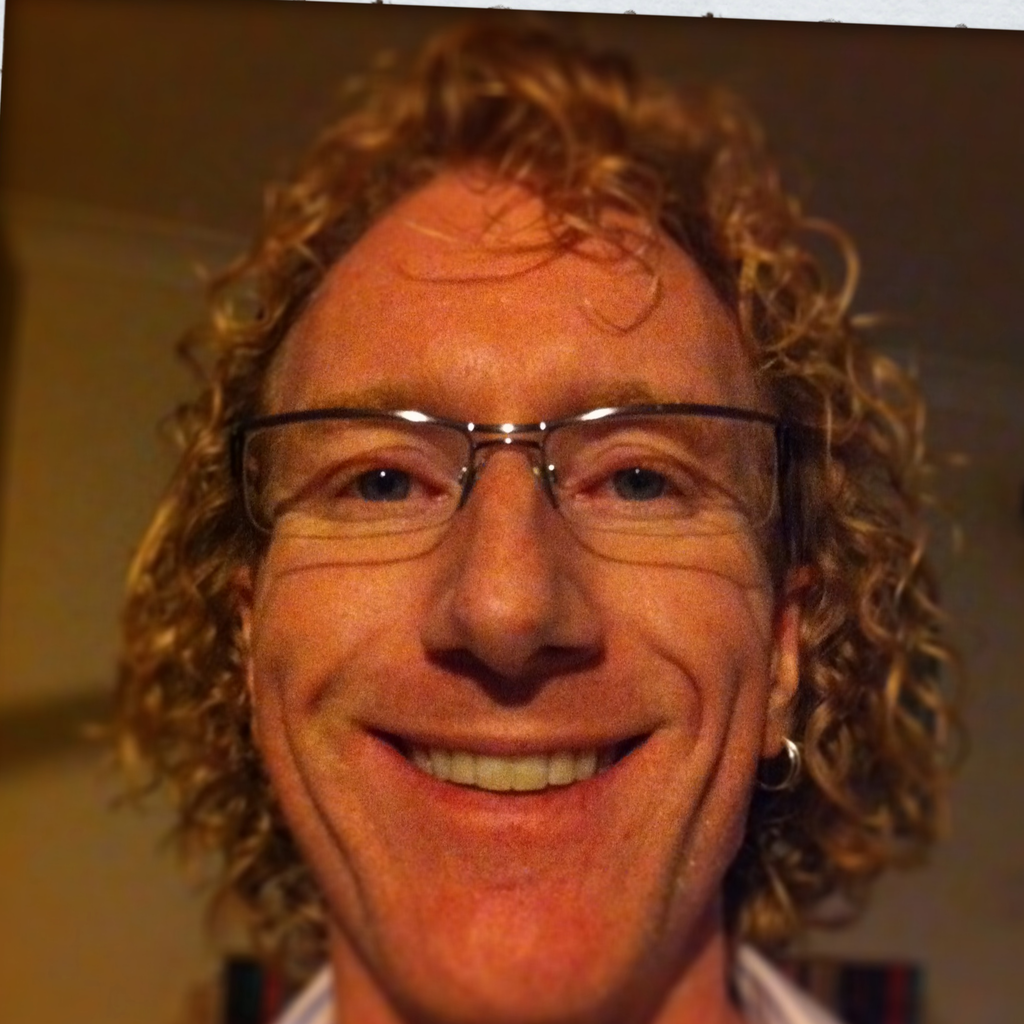}}
    \subfigure{\includegraphics[scale=0.08]{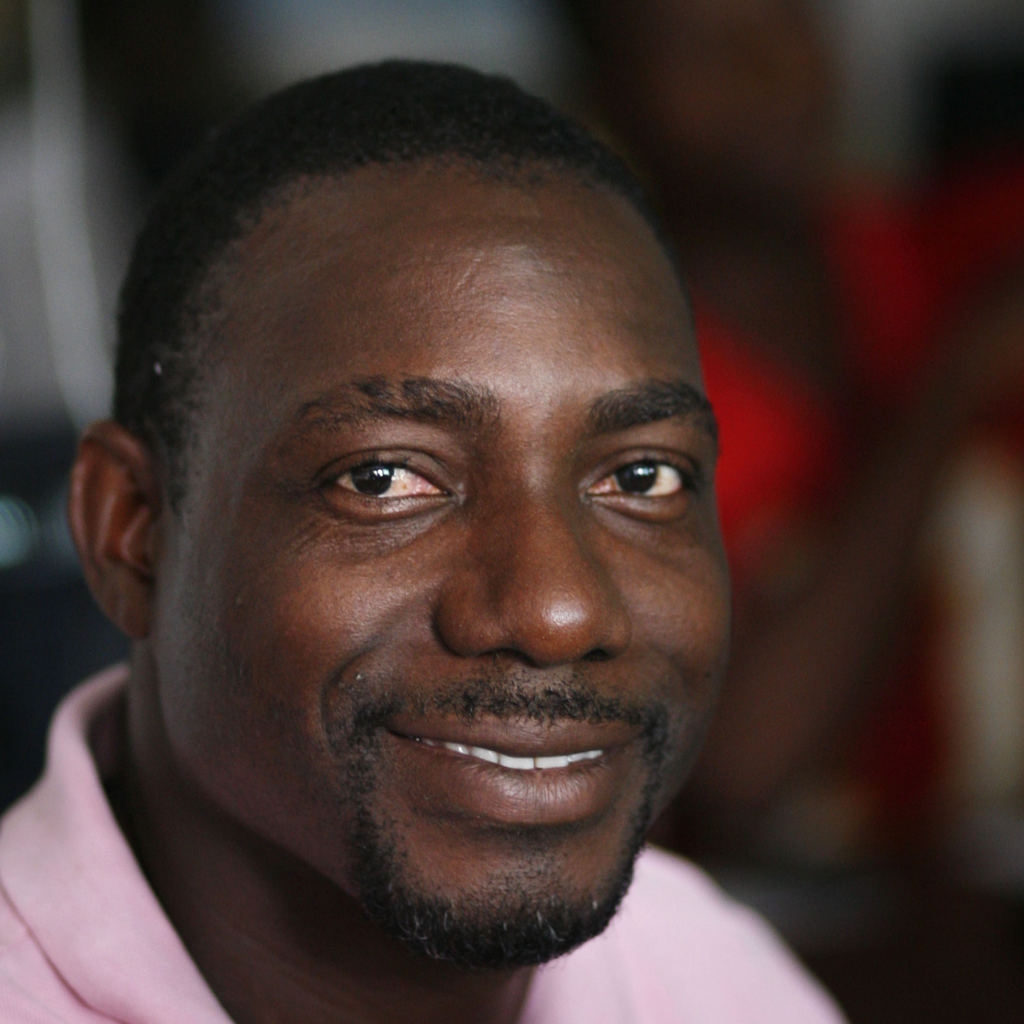}}
    \caption{Some digital images considered from the work of Karras \textit{et al.} \cite{karrasetal:2019} to build our dataset of printed images.}
    \label{fig:examples}
\end{figure}

To select the images to print in the first dataset, we choose images from a dataset that has particular importance in the digital image forensics literature. These images come from the original subset of human faces from the Flickr-Faces-HQ (FFHQ) dataset \cite{karrasetal:2019}. We use these images for two reasons: (i) they have enough samples to generate a large dataset of printed images, which can be used by data-hungry techniques such as those based on deep learning; and (ii) they can be used to develop methods focusing on applications (\textit{e.g.}, child pornography) for which printing patterns usually found in other datasets (\textit{e.g.,} barcodes and text) are not useful. Some examples of the images included in the first section of the dataset are shown in Figure \ref{fig:examples}.

We choose to print the images in the dataset with printers that are diverse enough to make the source attribution problem challenging enough for state-of-the-art techniques. The initial version\footnote{As we said, we are planning to continuously update the dataset with new images, printed with other printers.} of such sub-dataset for source attribution contains 1,600 printings from the printers listed in Table \ref{printers-list}. We would like to highlight the difficulties associated with such a dataset as it contains modern printers, with some of them being professional laser printers that were commercialized in the last five years. The dataset also contains printers with different printing resolutions: for example, printers \#1 and \#8 have native resolutions different from the others (600 $\times$ 600 dpi).

\begin{table}[h!]
    \centering
    \includegraphics[scale=0.65]{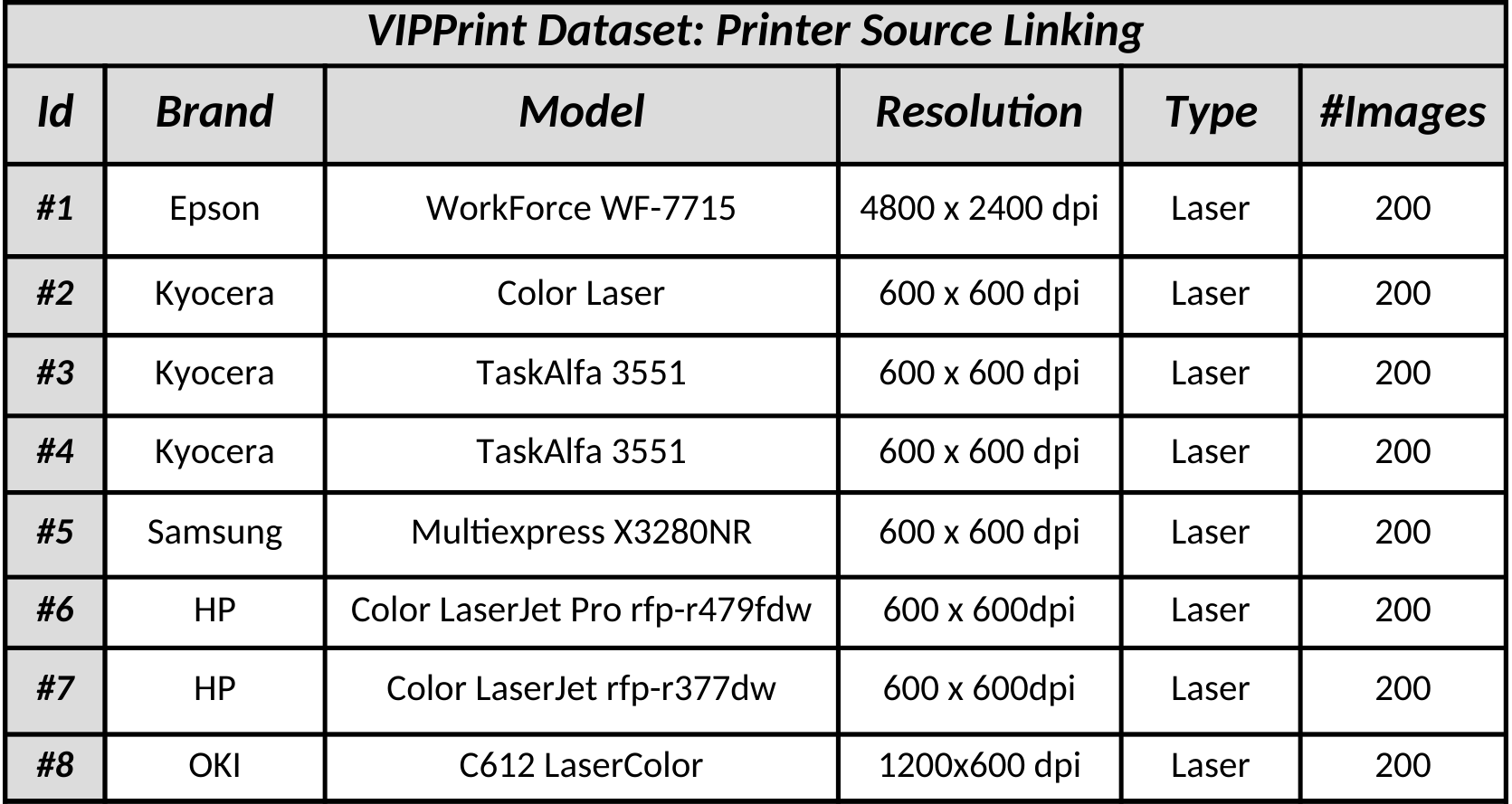}
    \caption{List of the eight laser printers that compose the first version of the VIPPrint dataset.}
\label{printers-list}
\end{table}

To scan the printed images, we used a scanner from the Kyocera TaskAlfa3551ci multifunctional printer (printer \#3 in Table \ref{printers-list}), with 600 $\times$ 600 dpi scanning resolution at the highest possible sharpness. The images are saved in a lossless compression configuration. As shown in Table \ref{printers-list}, we printed 200 images per printer. For that, we used 50 A4 sheets of paper, printing four images per sheet using the landscape orientation and then extracting individual patches. 

\begin{figure}[h!]
    \centering
    \includegraphics[scale=0.4]{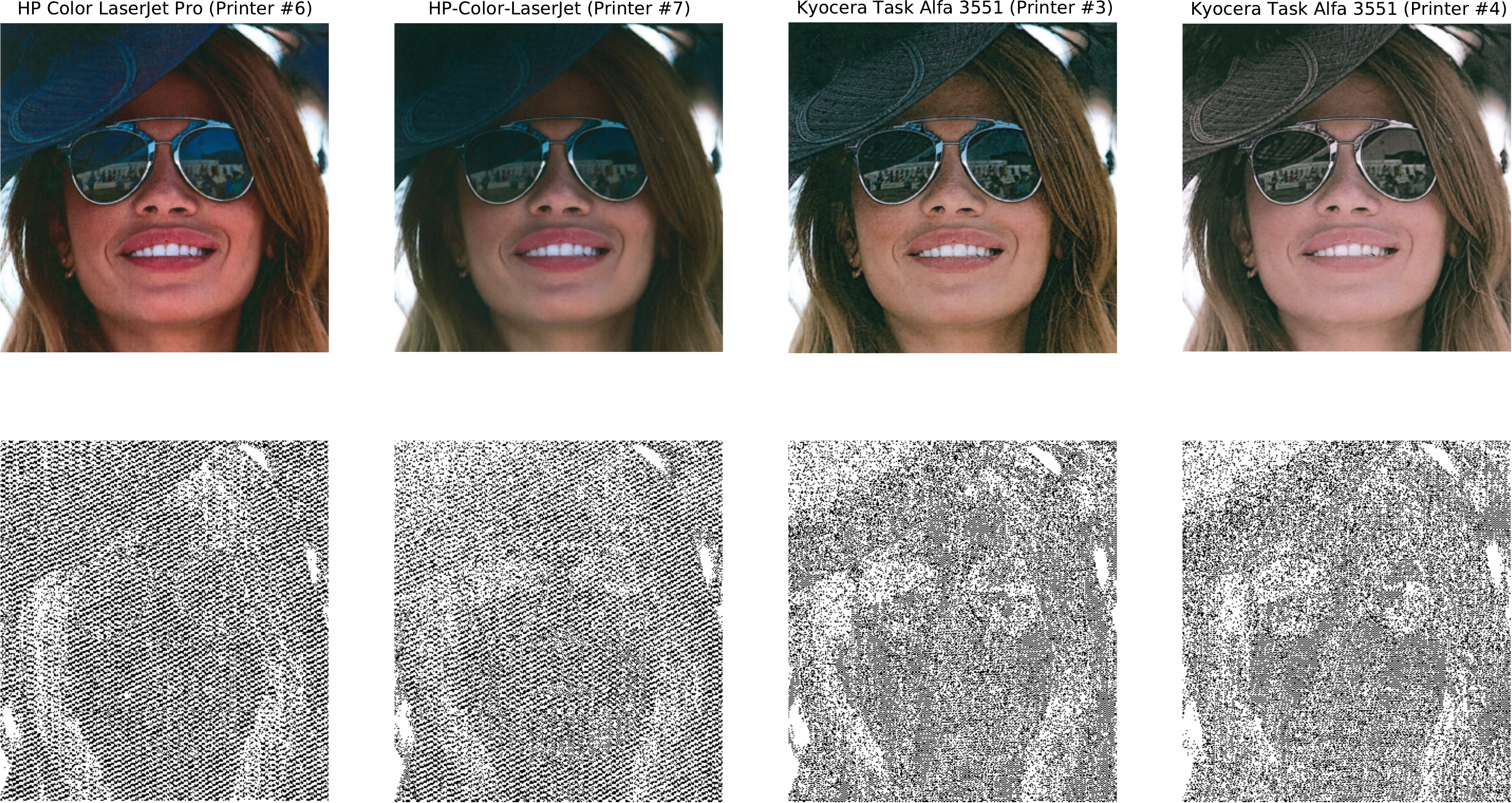}
    \caption{The same image (193.jpg) printed by four different printers and their corresponding HH Discrete Wavelet transform subbands (luminance component).}
    \label{fig:diff_wavelet}
\end{figure}

To illustrate the difficulties associated with source-linking of the images in the dataset, in Figure \ref{fig:diff_wavelet} we show the same image printed by different printers and its HH DWT subbands, which were used by \textit{Choi et al.} \cite{choietal:2009} to perform source attribution of colored documents. Very subtle differences can be seen in HH subbands of different printers from the same brand but different models (Printers \#6 and \#7 in Figure \ref{fig:diff_wavelet}), but no clear differences in the HH subband when using the same brand and model (Printers \#3 and \#4 in Figure \ref{fig:diff_wavelet}).

As we are aware that 200 images per printer may not be enough for data-hungry techniques such as those based on deep learning, we produced a second set of images containing Regions of Interest (ROI) extracted from the \textit{full images} set. The importance of the ROI sub-dataset for classification algorithms is three-fold: (i) it may filter only areas that are useful for recognition (\textit{e.g,} areas containing edges); (ii) such areas can be input to techniques that require lots of data such as data-driven approaches; and (iii) they allow the classification of documents through the fusion of their ROIs classification, providing the most accurate results. Such a strategy was validated several times before in the digital forensics domain, such as in works for camera source attribution \cite{bondietal:2017,ferreiraetal:2018}, anti-spoofing solutions \cite{agarwaletal:2018} and other works in laser printer source attribution \cite{anselmoetal:2015,bibietal:2019}.

To extract the ROI patches, we used an approach similar to that adopted in \cite{agarwaletal:2018} to tackle rebroadcast attacks in a data-driven classification scenario. In particular, we extract image patches by firstly applying Canny filtering to the whole input image, and then dividing the edge image into squared blocks of varying sizes. Then, we calculate the energy $E$ of the image patches using the horizontal ($H$), vertical($V$), and diagonal ($D$) sub-bands of the Discrete Wavelet Transform as follows:

\begin{equation}
E=\dfrac{\sum_{i=1}^{N} \sum_{j=1}^{N} H(i,j)^2 + \sum_{i=1}^{N} \sum_{j=1}^{N} V(i,j)^2 + \sum_{i=1}^{N} \sum_{j=1}^{N} D(i,j)^2 }{M^2},    
\end{equation}
where N is the number of values in the sub-bands of DWT and M is the fixed size of the squared patches.

\begin{figure}[ht!]
    \centering
    \includegraphics[scale=0.55]{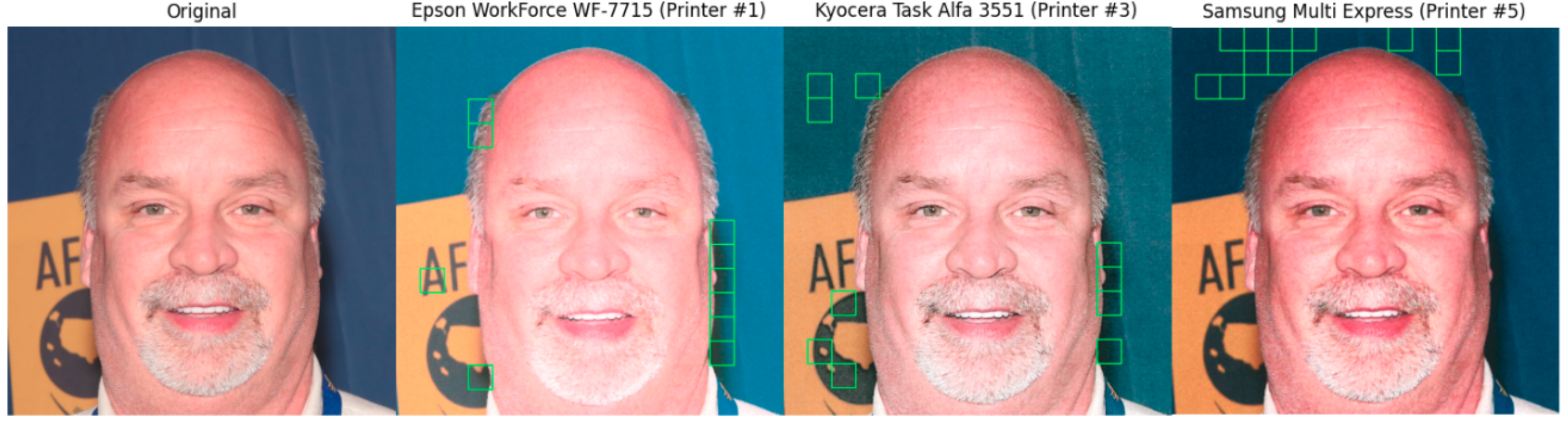}
    \caption{Image 139.jpg of the dataset (first column) and top-10 energy blocks of size 128 $\times$ 128 for different versions of the image printed by various printers (remaining columns).}
    \label{fig:patches-139.jpg}
\end{figure}

After such a metric is calculated for each patch, we ranked the image patches according to their $E$ and selected the top-10 energy patches per image. The patches selected in this way compose the RoI subdataset. We chose to focus on patches with the highest energy in the DWT sub-bands of edge images because such sub-bands contain useful information about edges in different directions, therefore the areas with high energy contain edge information to be used by printer attribution techniques focusing on banding artifacts.  For this second set of images, we choose patch sizes of $28 \times 28$,  $32 \times 32$, $64 \times 64$, $128 \times 128$, $224 \times 224$, $227 \times 227$, $256 \times 256$ and $299 \times 299$ to make such images suitable to most of the deep learning approaches available today. The RoI subdataset contains, therefore, 128,000 high energy patches. Figure \ref{fig:patches-139.jpg} shows some example of high energy patches selected according to the proposed criterion.     

\subsection{VIPPrint Dataset for synthetic GAN Images Detection}
\label{deepfake-dataset}

Detecting if an image is a \textit{deepfake}, \textit{i.e.}, if it has been artificially generated by a GAN, is an increasingly trendy topic in multimedia forensics. In the context of a criminal investigation, for instance, assessing that an image has been taken by a digital camera rather than having been generated artificially can be of fundamental importance to assess the trustfulness of a proof. As another example, in a social media scenario, detecting synthetic images may be useful to understand that a misinformation campaign supported by fake media is ongoing.

So far, research in this area has focused on digital documents, as they are intrinsically linked to fake news in social media. Several strategies have been proposed to deal with such a problem, including analysing the co-occurrence behavior of pixels in RGB channels \cite{lakshmananetal:2019}, cross-spectral co-occurrence between pairs of RGB channels \cite{barnietal:2020}, discrepancies in color spaces \cite{haodongetal:2020}, contrastive loss between original and fake images \cite{hsuetal:2018} and also other variations of deep learning approaches \cite{marraetal:2018,bonettinietal:2020}. On the contrary, very few works have considered the detection of deepfake printed images. To date and to the best of our knowledge, the only approach available to deal with the detection of printed manipulated images focuses on the identification of simple manipulations such as Gaussian blurring, Median filtering, resizing and JPEG compression \cite{jamesetal:2020}. Yet, printing and scanning back deepfake images is one of the easiest and most effective ways to fool media forensic techniques thought to work in the digital domain.

To promote further research on this topic, we built a second section of the VIPPrint dataset, containing a very large number of natural and GAN-generated face images. Specifically, we printed and scanned a total of 40,000 face images using a Kyocera TaskAlfa3551ci (Printer \#3 in Table \ref{printers-list}) in the following configurations:

\begin{itemize}
\item 16,000 pristine and 16,000 fake images generated by StyleGAN2 \cite{karrasetal:2019}.
\item 3,500 pristine and 3,500 fake images generated by ProgressiveGAN \cite{karrasetal:2017}.
\item 500 pristine and 500 fake images generated by StarGAN \cite{choietal:2017}.
\end{itemize}

\begin{figure}[h!]
    \centering
    \subfigure[Digital]{\includegraphics[scale=1.1]{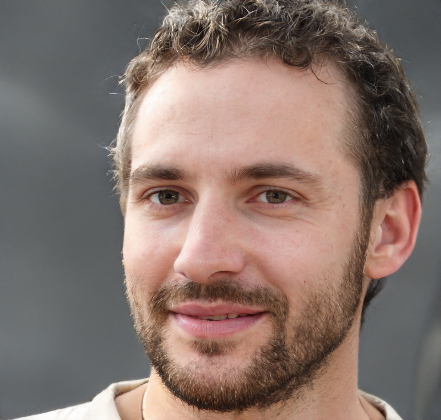}}
    \subfigure[Printed]{\includegraphics[scale=0.487]{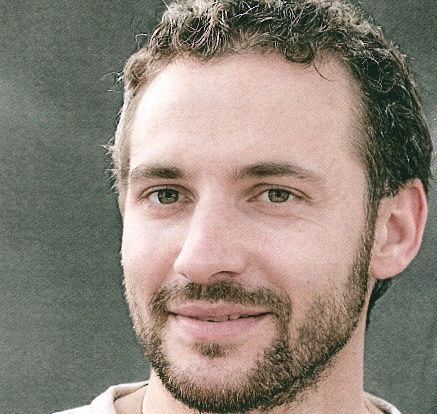}}

    \caption{A StyleGAN2 generated image in its original and printed-scanned versions.}
    \label{fig:GanDifferencesPrintedScanned}
\end{figure}

\begin{figure}[h!]
    \centering
    \subfigure[Digital Zoomed]{\includegraphics[scale=0.6]{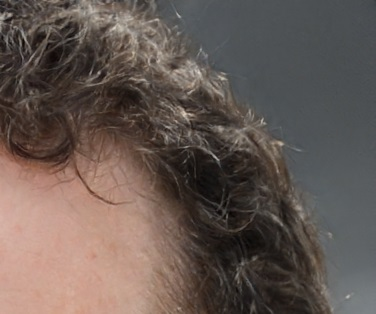}}
    \subfigure[Printed Zoomed]{\includegraphics[scale=0.481]{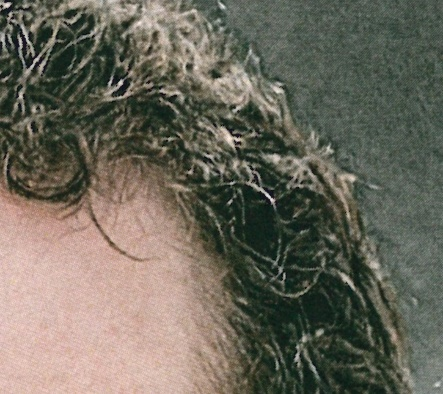}}
    \caption{Zoomed regions of the same pictures in Figure \ref{fig:GanDifferencesPrintedScanned}}
    \label{fig:GanDifferencesPrintedScannedPatches}
\end{figure}

\begin{figure}[htbp!]
\centering
    \subfigure[Digital Co-occurrence Matrix (R)]{\includegraphics[scale=0.37]{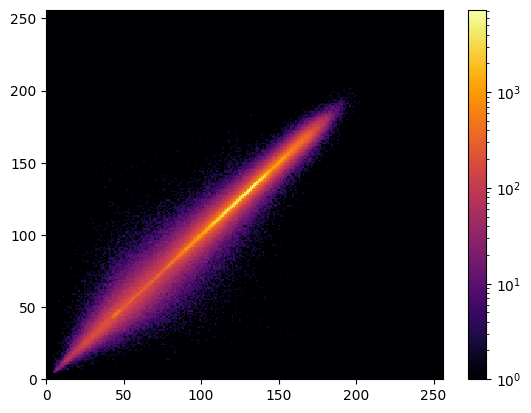}}
    \subfigure[Digital Co-occurrence Matrix (G)]{\includegraphics[scale=0.37]{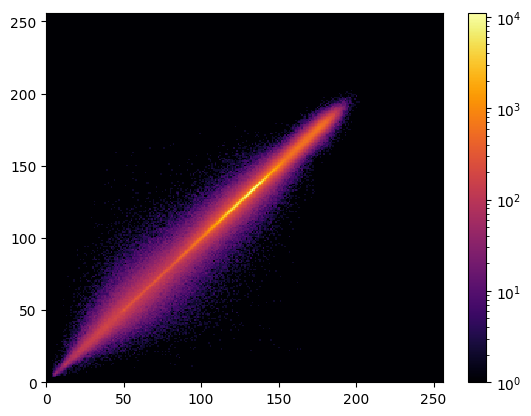}}
    \subfigure[Digital Co-occurrence Matrix (B)]{\includegraphics[scale=0.37]{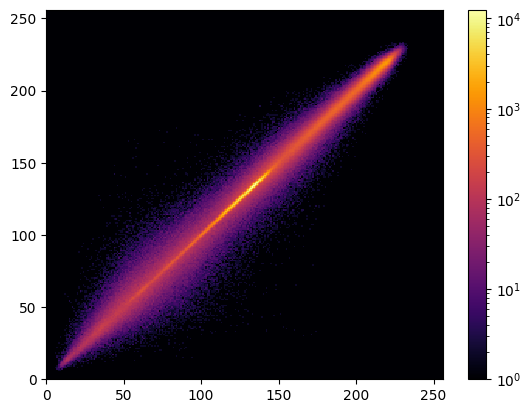}}
    \hfill
    \subfigure[Co-occurrence Matrix after print and scan (R)]{\includegraphics[scale=0.37]{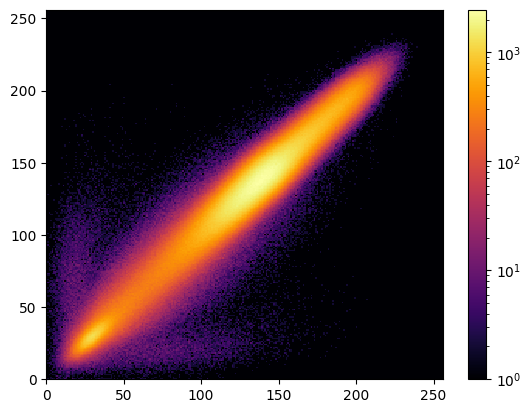}}
    \subfigure[Co-occurrence Matrix after print and scan (G)]{\includegraphics[scale=0.37]{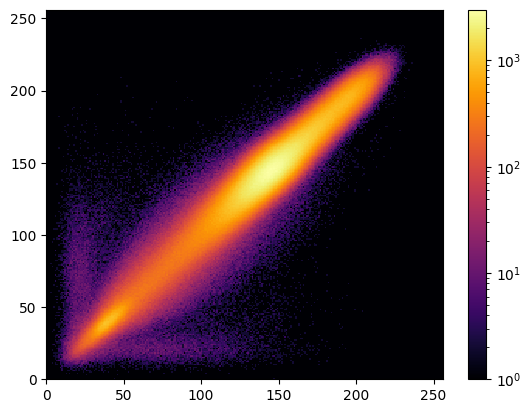}}
    \subfigure[Co-occurrence Matrix after print and scan (B)]{\includegraphics[scale=0.37]{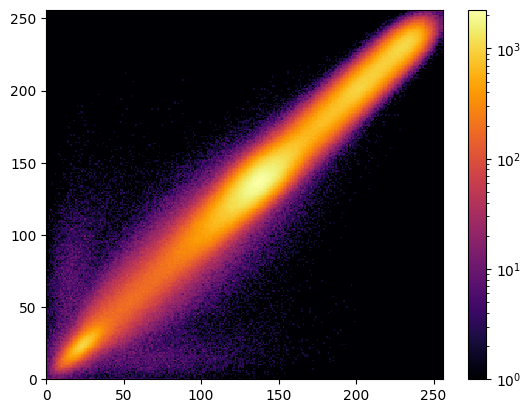}}
    \caption{Co-occurrence matrices proposed in \cite{lakshmananetal:2019} to discriminate GAN generated images from natural ones and their behavior in digital and Print and Scan images: (top) GAN image co-occurrence matrices in the digital format (bottom) GAN image co-occurrence matrices after Print and Scan.}
    \label{fig:co-occurence-rgb}
\end{figure}

The first difficulty with these images is the heavy distortion introduced in pixels after printing and scanning. Figure \ref{fig:GanDifferencesPrintedScanned} shows how a GAN image is degraded after printing and scanning. The calculated Structural Similarity Index \cite{ssi} of such images is 0.41 and the Peak Noise to Signal Ratio is 17.65 dB, which corresponds to intense image degradation. The noisy texture of the degradation is visible in the zoomed regions of the digital and printed images highlighted in Figure \ref{fig:GanDifferencesPrintedScannedPatches}. It is pretty clear from the analysis of this picture that distinguishing between printed pristine and GAN images by looking at textural artifacts only is an extremely difficult task. To further substantiate this hypothesis, in Figure \ref{fig:co-occurence-rgb} we show the co-occurrence matrices of the RGB bands before and after scanning and printing. The change between the matrices is dramatic, as the image is basically rebroadcasted by another image generation device (\textit{i.e.}, a scanner), possibly erasing the artifacts used to distinguish between natural and GAN images.

%

\begin{figure}[ht!]
    \centering
    \includegraphics[scale=0.54]{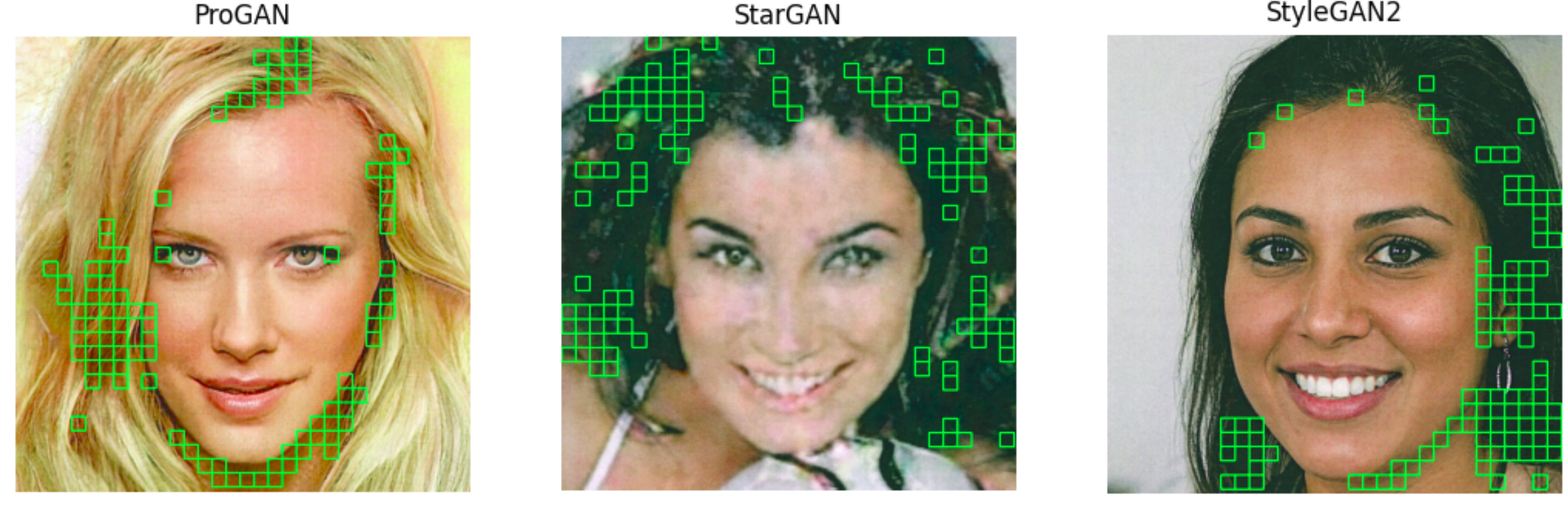}
    \caption{Sample printed pictures from each of GANs in our dataset with their 64 $\times$ 64 top-100 energy blocks.}
    \label{fig:patches-gan}
\end{figure}

As for the source attribution dataset, we also built a ROI dataset by applying high energy patch extraction and ranking. However, for this specific problem, the top-100 energy patches are selected\footnote{Depending on the size of the patches, the selection may correspond to selecting all the patches with non-zero energy.}. This new subset contains, for the StyleGAN2 case 1,109.822 patches for the size $299 \times 299$, and 2,392.469 patches for the $224 \times 224$ size. Patches for other dimensions and GANs can also be extracted by following the same approach. In the rest of the paper, we will focus on StyleGAN2 images, since they are by far the most difficult to discriminate. Figure \ref{fig:patches-gan} shows an example of some GAN images of our dataset along with the selected patches.\\

%% file: experimental-setup.tex
\section{Experimental Setup}
\label{experimental-setup}

In this section, we discuss the experimental setup we used to assess the difficulties associated with source attribution (a multiclass classification problem) and GAN image detection (a binary classification problem) on the images of the VIPPrint dataset. Specifically, we present the metrics used for the experiments, the experimental methodology, the statistical tests we adopted (when applicable), and the baseline approaches we tested together with their implementation details.

\subsection{Metrics}

Even if authentication and source linking are different classification problems (\textit{i.e.,} a binary and a multi-class problem respectively), the performance achieved by different methods on such tasks can be measured with similar metrics, by paying attention to interpret them properly according to the considered task. The set of metrics we have used is described in the following.

\subsubsection{Recall}

For binary classification problems, the \textit{recall}, also known as true positive rate, indicates the percentage of correctly classified positive samples and is calculated as
\begin{equation}
\centering
Recall = \frac{TP}{TP+FN},
\end{equation}
where $TP$ (True Positives) represents the number of samples correctly classified as positives, and $FN$ (False Negatives) is the number of positive samples wrongly labeled as negative. In our binary classification problem, the Recall metric measures how many GAN images in the testing set were correctly detected as such.

For the multiclass problem of source attribution, a similar metric can be used, with the difference of considering the recall for each class (\textit{i.e.,} for each printer label) from a confusion matrix $M$, which disposes the number of predicted samples in all the classes. Considering that each entry $M_{i,j}$ in the confusion matrix gives the number of predictions for class $i$ when the real class is $j$ (with correct classifications located at $M_{i,i}$), the recall for a class $i$ is the total number of correctly classified samples of class $i$ in the confusion matrix $M$, divided by the total number of samples in class $i$, that is:

\begin{equation}
    Recall_i=\frac{M_{i,i}}{\sum_j M_{j,i}}.
\end{equation}


We used the weighted approach to calculate the recall, so the recall reported is the mean of recalls for all $i$ classes weighted by the number of true instances for each class, that is:

\begin{equation}
    Recall=\frac{\sum_{i=1}^{N} Recall_i \times c_i }{\sum_{i=1}^{N} c_i},
\end{equation}
where $c_i$ is the cardinality (or the number of elements) of class $i$ in the testing set.

\subsubsection{Precision}

As a metric complementary to the Recall, we are interested in the classification precision, which is the fraction of correctly classified positives out of all the instances classified as such in a binary classification problem (in our case, GAN images detection). That is
\begin{equation}
\textit{Precision}=\displaystyle \frac{TP}{TP+FP}.
\label{precision}
\end{equation}

For the case of source attribution (a multiclass problem), we considered the per-class precision in a way similar to what we did for the recall. The precision for a class $i$ is the total number of correctly classified samples in $i$ out of all the assignments to class $i$ (correct or incorrect), that is:

\begin{equation}
    Precision_i=\frac{M_{i,i}}{\sum_j M_{i,j}}.
\end{equation}

We also report the average precision across all classes in the multiclass problem, by using again a weighted mean, or

\begin{equation}
    Precision=\frac{\sum_{i=1}^{N} Precision_i \times c_i }{\sum_{i=1}^{N} c_i}.
\end{equation}

\subsubsection{F-measure}

The most important metric for both problems is f-measure ($F$). It measures the harmonic mean of precision and recall and is calculated as follows for the binary classification case:
\vspace{-1.5pt}
\begin{equation}
\textit{F}= 2 \times \frac{\mbox{\textit{\textit{P}}} \times R}{\mbox{\textit{P}} + R}.
\label{fmeasure}
\end{equation}

For the multi-class source attribution problem, we calculate the f-measure individually for each class by using per-class precisions and recalls and weighting them over all classes, exactly as done for the precision and the recall. That means

\begin{equation}
    F=\frac{\sum_{i=1}^{N} F_i \times c_i }{\sum_{i=1}^{N} c_i}.
    \label{fmeasuremulti}
\end{equation}

\subsubsection{Accuracy}

As a final metric, we considered the accuracy. In a binary classification problem, it is defined as the total number of samples correctly classified (in both classes) divided by the number of samples under investigation:
\begin{equation}
\textit{Accuracy}= \frac{TP+TN}{TP+TN+FP+FN}.
\label{accuracy}
\end{equation}

In the multiclass problem, we simply have:

\begin{equation}
    Accuracy_i=\frac{TP_i+TN_i}{TP_i+TN_i+FP_i+FN_i},
\end{equation}
where $TP_i$ is the correct number of samples being classified as the class $i$, $TN_i$ is the number of samples correctly classified as not being from class $i$, $FN_i$ is the number of elements from class $i$ wrongly classified as being from another class, and $FP_i$ is the number of elements from another class wrongly classified as being from class $i$. The accuracy reported is the weighted average of all the accuracies for all classes. That means

\begin{equation}
    Accuracy=\frac{\sum_{i=1}^{N} Accuracy_i \times c_i }{\sum_{i=1}^{N} c_i}.
\end{equation}

\subsection{Experimental Methodology}
\label{methodology}

To validate the experiments carried out on the VIPPrint dataset we followed two different approaches, depending on which application we are considering. In the source attribution scenario, we choose the $5 \times 2$ cross-validation protocol, as it is considered an optimal benchmarking protocol for machine learning algorithms \cite{Dietterich:98} and was also used in other works on printer attribution \cite{anselmoetal:2015, anselmoetal:2017}. According to such a protocol, five iterations of twofold cross-validation are carried out. In other words, the data is firstly randomized, and 50\% of data is selected as the training set with the other 50\% is used for testing. Then the process is inverted. As stated before, this process is repeated five times (five rounds), resulting in ten experiments of training and testing the machine learning classifiers. Additionally, when using deep learning approaches, we also need validation data in order to help training. Therefore, we further split the 50\% of training data into training data and validation data, with a ratio equal to 70:30.  

It is important to notice that, in contrast to camera source attribution validation approaches commonly used in the literature \cite{bondietal:2017,ferreiraetal:2018} that use totally random images generated by different cameras, for source attribution of printed documents the same document can be printed by different printers \cite{anselmoetal:2015,anselmoetal:2017}. In this paper, we consider the source attribution problem as a \textit{closed set} multiclass problem, where we classify documents printed by known printers in our dataset.

For the GAN-image detection task, we took a set of detectors and trained them on the original digital images, as done in the original papers, and assessed their performance on printed and scanned images. We focus on the detection of the StyleGAN2 images in the VIPPrint dataset, as they are by far the best quality GAN images in the dataset. The procedure we have followed to evaluate the performance of the detectors is a simple one: we use 24,000 digital images for training, 6,000 digital images for validation and then we used 2,000 printed and scanned images from our dataset for testing the detectors. All the sets are independent and stratified (\textit{i.e.}, images in one set are not present in the others and there is an equal number of images per class). 

\subsection{Statistical Tests}

To verify that the source attribution results are statistically significant, we perform a series of two tests in the $5 \times 2$ cross-validation procedure. The first one, which we call a \textit{pre-test}, is used to confirm that all the techniques considered in the experiment are statistically different. If they pass this test, then we do a \textit{post test} that compares the results in a pairwise manner. The pre-test is done in the distributions of f-measures calculated from Equation \ref{fmeasuremulti} at ten runs of the $5 \times 2$ cross-validation experiments for each technique. The test is applied to an input matrix of $n$ rows (where $n$ is the number of tested approaches) and ten columns, which are the ten f-measures resulting from the 10 runs. The test aims at verifying if the distributions of all the sets of F-measures change significantly. We use the Friedmann test \cite{friedman:1937} for this first step, with a confidence level of 95\%. In other words, if the calculated p-value is below 0.05, then the null hypothesis, which says that there is no statistically significant difference between the F-measures distributions, is rejected and we can pass to the next test. 

For the \textit{post test}, which tests the statistical relevance of each pair of approaches, we consider the Student's t-test \cite{student08ttest}. This test can determine if there is a significant difference between the means of F-measures distributions taken pairwise. To apply this test to our scenario, we also consider the same set of $5 \times 2$ F-measures results, but now for each possible pairs of approaches. In this test, we set again the confidence level to 95\%: if the calculated p-value is below 0.05, then the null hypothesis, which states that there is no statistical significance between the performance of the pair of approaches, is rejected.

\subsection{Baseline methods}

In this section, we briefly describe the baseline methods considered in our tests.

\subsubsection{Source attribution}

For this problem, we select 12 approaches divided into three sets. In the first set, which we call \textit{image texture descriptors}, we used a set of common descriptors that are mainly used for image characterization. For the source attribution task, such descriptors can be useful to differentiate printers banding artifacts efficiently if the analyzed patterns do not change much and, therefore, they normally exhibit good performance in some specific printer source attribution tasks \cite{anselmoetal:2015,anselmoetal:2017,joshietal:2018,joshikhanna:2020}. We considered four approaches in this set as follows.
\begin{itemize}
    \item The Gray Histogram \cite{jainvailaya:1996} (hereafter referred to as \texttt{GH}), which divides the grayscale version of the analyzed image into a fixed number of blocks. Then, a histogram of gray intensities is calculated for each block and all the histograms together are used to generate a description vector.
    
    \item The Histogram of Oriented Gradients \cite{dalatriggs:2005} (hereafter referred to as \texttt{HOG}), which extracts the edges in the image by means of the Sobel kernel gradients, then it computes the gradient for all the orientations. Finally, a histogram of such orientations is fed at the input of a machine learning classifier.

    \item The Edge Histogram \cite{jainvailaya:1996} (hereafter referred to as \texttt{EH}) is similar to \texttt{HOG}. However, it calculates, for each block, the dominant edge orientation instead of all of them, and the descriptor is a histogram of these orientations.
    
    \item The Local Binary Patterns \cite{Ojalaetal:96} divides the image into blocks and compares each pixel in a block to all its neighbors. If the pixel in the center of the block is greater than a neighbor's value, then a 0 digit is written (1 otherwise). Considering eight neighbors in each block, 8-digit binary numbers are generated for each pixel in a block. Such digits are converted to decimals and histograms for each block are calculated, normalized, and concatenated to describe the image.
\end{itemize}

The second class of approaches have already been introduced in section \ref{related-work}, and are referred to as \textit{Feature Based Source Printer Source Attribution Baseline Techniques}. These approaches have been already validated in the printer source attribution problem by previous works in the literature, and they are: 

\begin{itemize}
    \item The multidirectional version of Gray level Co-occurrence matrix (\texttt{GLCM-MD}) from Ferreira \textit{et al.} \cite{anselmoetal:2015}.
    \item The multidirectional and multiscale version of the same approach proposed in \cite{anselmoetal:2015} (\texttt{GLCM-MD-MS}).
    \item The Convolutional Texture Gradient Filter in a $3\times3$ window \cite{anselmoetal:2015} (\texttt{CTGF-3X3}).
    \item The 39 statistical features from the diagonal Discrete Wavelet Transform sub-band from Choi \textit{et al.} \cite{choietal:2009} (\texttt{DWT-STATS}).
\end{itemize}


Finally, the third set of approaches belong to the class of \textit{Data-driven Baselines} and rely on the training of deep neural networks. For this set, we considered several convolutional neural network approaches analyzed in \cite{bibietal:2019} for printer source attribution. These are:

\begin{itemize}
    \item The 16 and 19 layers version of the VGG convolutional neural network \cite{Simonyan15} (\texttt{VGG-16} and \texttt{VGG-19}).
    \item The 50 and 101 layers versions of the RESNET convolutional neural networks \cite{heetal:2016} (\texttt{RESNET-50} and \texttt{RESNET-101}). 
\end{itemize}

\subsection{Printed and Scanned GAN image detection}

For the deepfake detection task, we choose a set of deep learning classifiers proposed in the literature for digital images. The first three approaches are based on ImageNet dataset pre-trained models and their use for GAN images detection was validated in the work of Marra \textit{et al.} \cite{marraetal:2018}. They are: 

\begin{itemize}
    \item The Densely connected networks \cite{HUANGETAL:2017} (\texttt{DENSENET})
    \item The third version of InceptionNet \cite{Szegedyetal:2016} (\texttt{INCEPTION-V3});
    \item The InceptionNet evolution considering fully separable filters \cite{chollet:2017} (\texttt{XCEPTION})
\end{itemize}
The other set of deep neural networks are \textit{ad-hoc} networks designed for the GAN detection problem. These networks act on pre-processed data, namely the co-occurency matrices of image channels, and they are: 

\begin{itemize}
    \item A CNN that acts on three co-occurence intra-channel matrices \cite{lakshmananetal:2019} (\texttt{CONET});
    \item A CNN that acts on six co-occurence matrices considering both intra- and inter-channel co-occurrences \cite{barnietal:2020} (\texttt{CROSSCONET}).
\end{itemize} 

All these five CNNs have been retrained on StyleGAN2 and pristine digital images as described in Section \ref{methodology}. In conclusion, we considered 12 baseline methods for the source attribution tasks and 5 for the GAN-image detection task. 

\subsection{Implementation Details}

To ensure the reproducibility of our results, we provide all the implementation details we used to achieve our results. We start with the source attribution approaches that we had to re-train from scratch. We had to do that because the eight printers used to build the VIPPrint dataset had never been used before in a source attribution problem. Then, we report the implementation details of the pre-trained baseline models on digital we used to distinguish printed and scanned GAN and pristine images.

We start with the feature engineering approaches. For \texttt{GH}, \texttt{LBP}, \texttt{EH}, \texttt{HOG} and \texttt{DWT} we used Python implementations, whereas for \texttt{GLCM-MD}, \texttt{GLCM-MD-MS} and \texttt{CTGF-3X3} we used Matlab implementations available at the authors' source code website \cite{ferreirasourcecode:2014}. Although the implementations use different programming languages, we used them only to extract the features, using a Linear SVM from Python's sci-kit-learn library\footnote{\url{http://scikit-learn.org}} for the final classification stage. We choose a Linear Kernel Support Vector Machines classifier as it is well suitable to deal efficiently with high dimensional features. We perform a grid-search approach to find the best parameters to train the classifiers for each of the 10 experiments. This is done by applying a five-fold cross-validation procedure to the training data only. The classifiers parameter $C$ is varied in the set $C=\{0.1, 1, 10, 100, 1000\}$ with the best value used to train the classifier.


In contrast to the previous approaches, those based on convolutional neural networks are applied patch-wise, with patches of size  $224\times224$ for \texttt{VGG-16}, \texttt{VGG-19}, \texttt{RESNET-50}, \texttt{RESNET-101}, \texttt{INCEPTION\_NET-V3} and \texttt{DENSENET}, and $299\times299$ for \texttt{XCEPTION\_NET}. The final classification result for an image is set to be the mode of the classifications obtained on single patches. Such approach is commonly known as \textit{Majority Voting} and was also validated in the printed document forensics research \cite{anselmoetal:2015,anselmoetal:2017,joshikhanna:2018,joshietal:2018,joshikhanna:2020}. To choose the patches, we applied the highest-energy procedure already described in section \ref{dataset}. The only exceptions to this rule are the \texttt{CONET} and \texttt{CROSS-CONET} networks for GAN image detection, which act on $256\times256$ co-occurence matrices computed on the entire images. We implemented these techniques by using Python's Tensorflow\footnote{\url{https://www.tensorflow.org/}} and Keras\footnote{\url{https://keras.io/}} libraries.

For a fair comparison of the data-driven approaches, we used the following common procedure to train the neural networks:

\begin{enumerate}
\item We fine-tuned the neural networks pre-trained on ImageNet with the input training data (\textit{e.g.,} high energized patches), by initializing the weights with Imagenet pre-trained weights. We tried other procedures such as fine-tuning only the top of the networks (\textit{i.e.}, the fully connected layers) and freezing the other layers, but the results were not worthwhile.

\item In the fine-tuning procedure, we cut off the top of these networks, replacing them with a layer of 512 fully connected neurons, followed by a 50\% Dropout layer and a final layer with eight or two neurons, depending on the task.

\item The networks are trained with the Steepest Gradient Descent optimizer \cite{Wardi1988}, with an initial learning rate of 0.01. The learning rate is reduced by a factor $\sqrt{0.1}$ once the validation loss stagnates after five epochs. We fix the learning rate lower bound to 0.5 $\times 10^{-6}$. We trained the networks on minibatches of size 32 for source attribution and 16 for GAN detection.

\item we set the maximum number of epochs for source attribution to 300 epochs. However, after 20 epochs we implemented an early stopping procedure if the validation loss does not improve. For deepfake detection, we chose 10 epochs and the early stopping condition is implemented after five epochs as we are using much more data. 

\item We used data augmentation for the source attribution task by using the following image processing operations: rotation, zoom, width shifts, height shifts, shears, and horizontal flips. For GAN detection, since much more training data is available (more than 300,000 images), we did not use any data augmentation.
\end{enumerate}

Finally, all the data presented in this paper, including the two datasets, the scripts for generating the high-energy blocks, $5\times2$ cross-validation data, and some of the source code used are all available at \url{https://bit.ly/3j1SQcM}. 


%% file: experiments.tex
\section{Experimental Results}
\label{experiments}

In this section, we discuss the results of our comparative study for both source attribution and GAN-image detection.

\subsection{Source Attribution}

An overall view of the average results we got for the 12 baseline source attribution techniques we have tested is reported in Table \ref{tab:results-printer}.

\begin{table}[h!]
    \centering
    \includegraphics[scale=0.89]{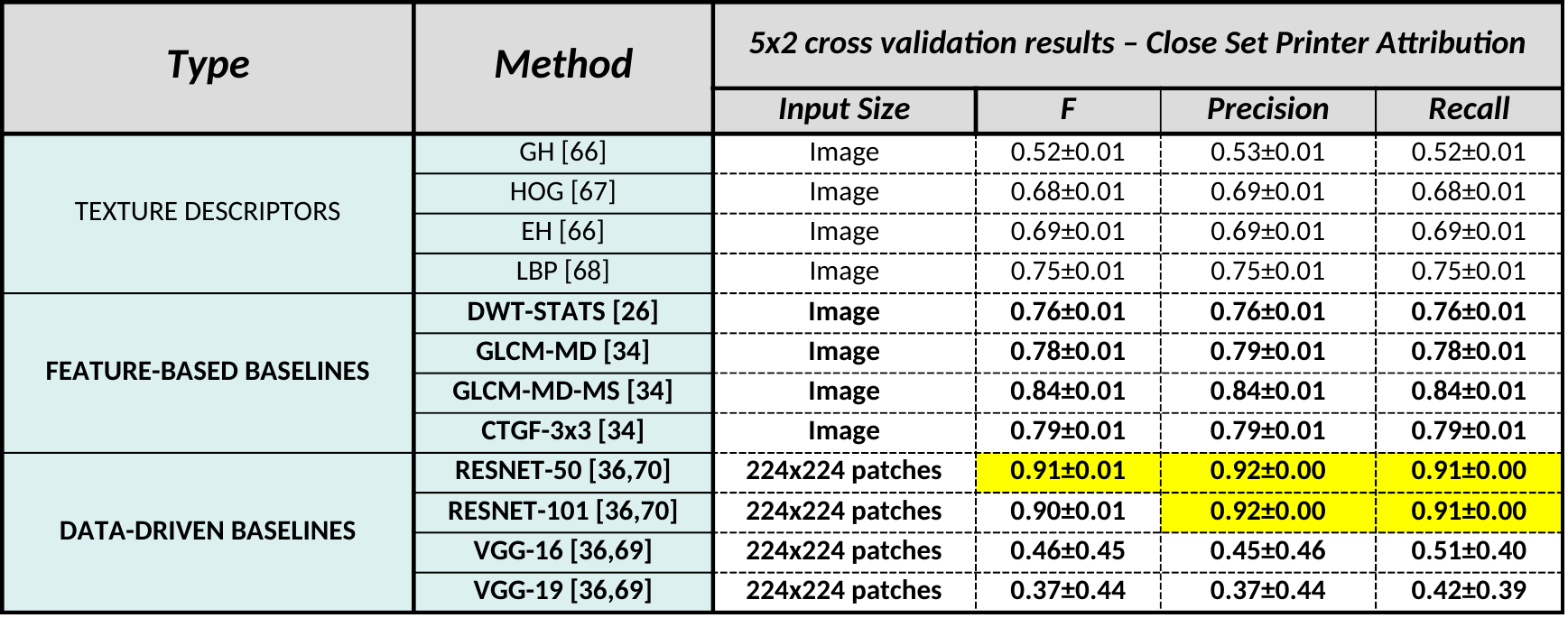}
    \caption{Average performance for the source attribution problem. The approaches are divided by category, with boldfaced entries denoting the solutions specifically designed for the source printer attribution problem. The best results for each metric are highlighted in yellow.}
    \label{tab:results-printer}\
\end{table}

The first aspect to be noticed in the results shown in Table \ref{tab:results-printer} is the bad performance obtained by methods based on general-purpose texture descriptors. The \texttt{GH} descriptor, for example, tries to discriminate printers by assuming that different printers print the same images using different colors, which is supposed to be seen in different histograms plotted in the $n$-dimensional space and clustered by hyperplanes such as from the SVM classifiers. Such assumption fails as the resulting f-measure (0.53) is pretty similar to a random guess. The approaches relying more on the effects of gradients and edges (\texttt{EH} and \texttt{HOG}), where the banding and other printing artifacts are more evident \cite{anselmoetal:2015}, achieve slightly better but still poor performance. The best f-measure in this class of techniques was obtained by the \texttt{LBP} descriptor ($F$ = 0.75). A possible explanation for the better performance of \texttt{LBP}  compared to other texture descriptors is that it explores gradient information by encoding, in several regions, the neighborhood relationships. This can better identify the behavior of printer patterns compared to other texture descriptors.

The second set of techniques includes approaches based on handcrafted features specifically tailored for the source attribution problem. To start with, we found that the performance of \texttt{DWT-STATS} ($F$ = 0.76) drops with respect to the performance reported in the original paper \cite{choietal:2009}, highlighting that different datasets with modern printers may confuse such characterization. Additionally, from the discussion done in Section \ref{dataset}, we found that considering statistics from a specific wavelet channel allows identifying different brands, but does not work well when identical devices are included in the set. Other descriptors from \cite{anselmoetal:2015} show better, but still unsatisfactory results. \texttt{CTGF-3X3} filters convolutional generated features, building their histogram in a gradient interval. Such an approach yields an average $F$=0.79, which is considered a good result when compared with the common texture descriptors we considered and discussed in the previous paragraph. We can also see from Table \ref{tab:results-printer} that better performances are also obtained by \texttt{GLCM-MD} ($F$=0.78) and \texttt{GLCM-MD-MS} ($F$=0.84). These approaches consider more directions in the neighborhood of pixels and more statistics in the co-occurrence matrices. Additionally, for \texttt{GLCM-MD-MS}, more scales are used in order to achieve invariance with respect to the size of the printed pattern. Such features can be considered ad-hoc texture features specific for laser printer attribution, achieving thus better performance than general texture descriptors. 

Finally, the last set of techniques are based on CNNs \cite{bibietal:2019}. Let us consider first the shallower networks, namely \texttt{VGG-16} and \texttt{VGG-19}. They provide the two worst results for all metrics, also showing a very high standard deviation indicating a very unstable training. Two possible reasons for such bad performance are the shallowness of the networks and their very simple architecture including only convolutional and pooling layers. Such an explanation is confirmed by the results got by deeper and more complex \texttt{RESNET-50} and \texttt{RESNET-101} CNNs. These networks exhibit (by far) the top-2 results of our tests, with $F$ = 0.91 for \texttt{RESNET-50} and $F$ =0.90 for \texttt{RESNET-101}.

To better investigate the differences between these networks, we start to discuss where they fail and succeed in the printer attribution task. Tables \ref{tab:cm-resnet50} and \ref{tab:cm-resnet101} show the confusion matrix of these approaches. It can be seen that both approaches have strong difficulties to discriminate two printers from Kyocera: the Color-Laser and a specific Taskalfa model (printers \#2 and \#3 of our dataset). This result is somewhat surprising because these printers are quite different physically (Taskalfa is a multifunctional printer and Color-Laser is an ordinary laser printer). One possible explanation is that the two printers could have shared some components in their manufacturing process. \texttt{RESNET-50} shows slightly better performance as it is less affected by such a problem and also because it classifies perfectly 4 out of 10 printers, instead of 3 out of 10.

\begin{table}[h!]
\centering
    \includegraphics[scale=0.89]{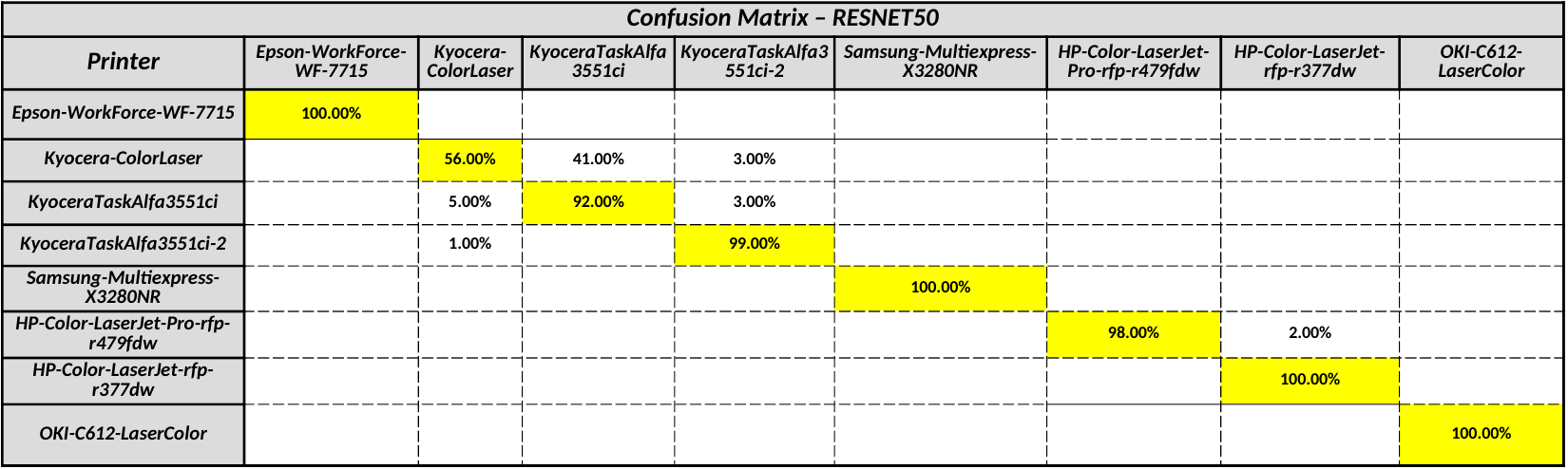}
    \caption{Confusion matrix of \texttt{RESNET-50} for the source attribution problem.}
    \label{tab:cm-resnet50}
\end{table}

\begin{table}[h!]
\centering

    \includegraphics[scale=0.89]{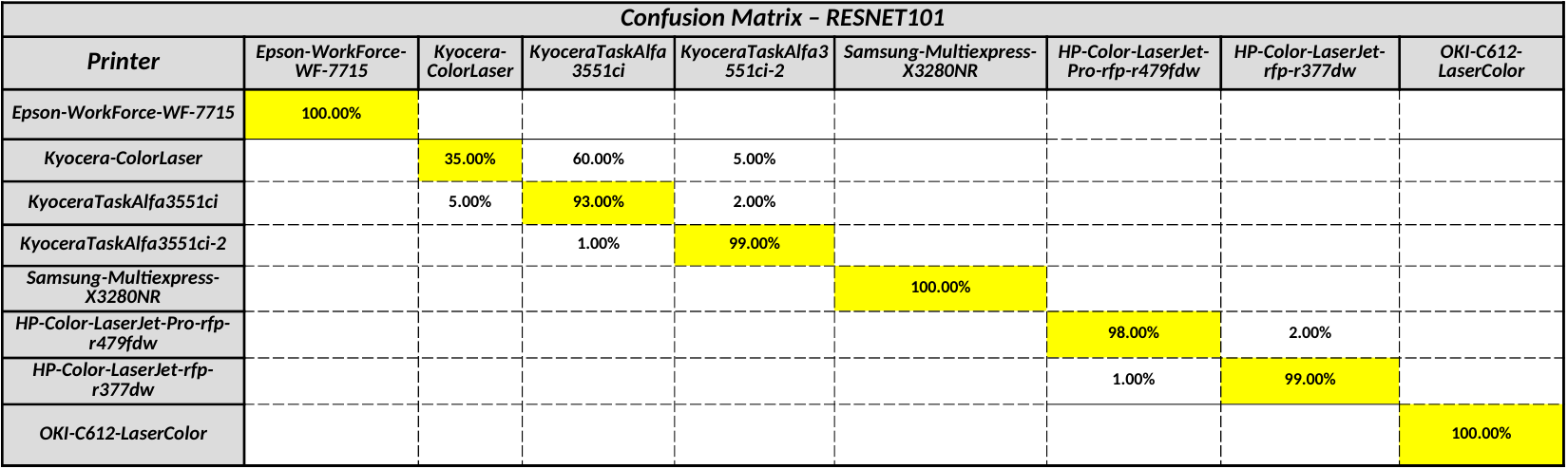}
    \caption{Confusion matrix of \texttt{RESNET-101} for the source attribution problem.}
    \label{tab:cm-resnet101}
\end{table}

As a final step of the printer attribution experiments, we analyze the statistical significance of the results. By applying the Friedmann test to 12 vectors (one for each approach) with the 10 f-measures, we got a p-value lower than $0.01$, thus proving that the differences in the f-measures between all the approaches are statistically significant. As a second step, the results of the pairwise statistical tests (Student's t-test) are shown in Table \ref{tab:tukey-results}.  

\begin{table}[h!]
\centering

    \includegraphics[scale=0.82]{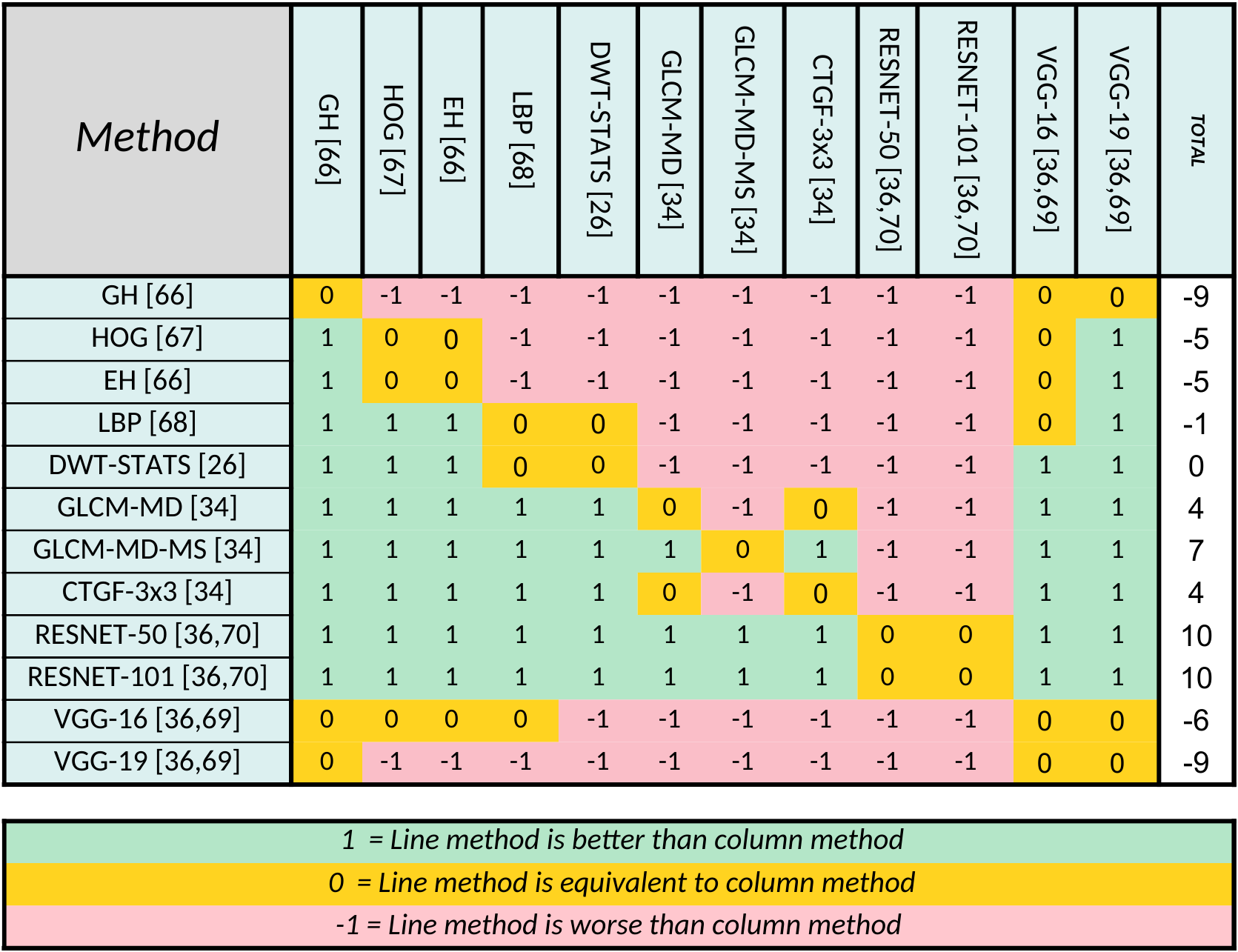}
    \caption{Pairwise comparison between different source attribution techniques.}
    \label{tab:tukey-results}
\end{table}

The first noticeable behaviour in Table \ref{tab:tukey-results} is that the large standard deviation of \texttt{VGG-16} does not allow to draw statistically significant conclusions for some of the comparisons, namely those with \texttt{GH}, \texttt{HOG}, \texttt{EH} and \texttt{LBP}.
Other cases where no statistically significant conclusions can be drawn are the comparison of \texttt{DWT-STATS} and \texttt{LBP}, and \texttt{HOG} with \texttt{EH}. 

Finally, we notice that the superior performance of \texttt{RESNET-50} and \texttt{RESNET-101} are confirmed by the results of the Student's t-tests with all the other methods. At the same time, the difference between the performance of these two networks is not statistically significant. Based on these observations, we can conclude that \texttt{RESNET-50} and \texttt{RESNET-101} represent the better solutions for the source attribution problem, even if their best performance, with an f-measure equal to 0.91 and the difficulties to distinguish some Kyocera printers, leave room for further improvements. 

\subsection{Detection of GAN images}

\begin{figure}[h!]
    \centering
    \subfigure[\texttt{CROSSCONET} \cite{barnietal:2020}]{\includegraphics[scale=0.3]{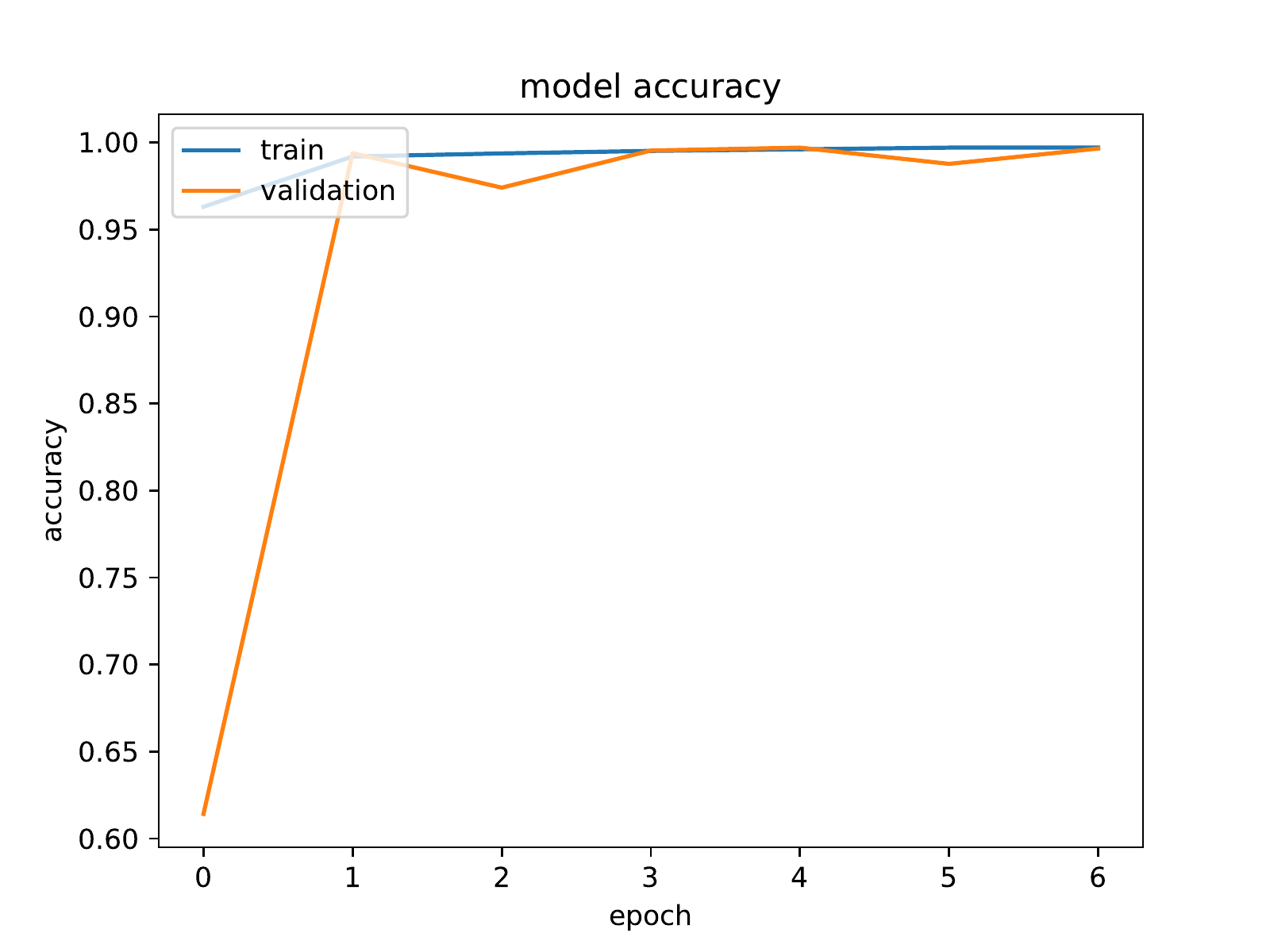}}
    \subfigure[\texttt{CONET} \cite{lakshmananetal:2019}]{\includegraphics[scale=0.3]{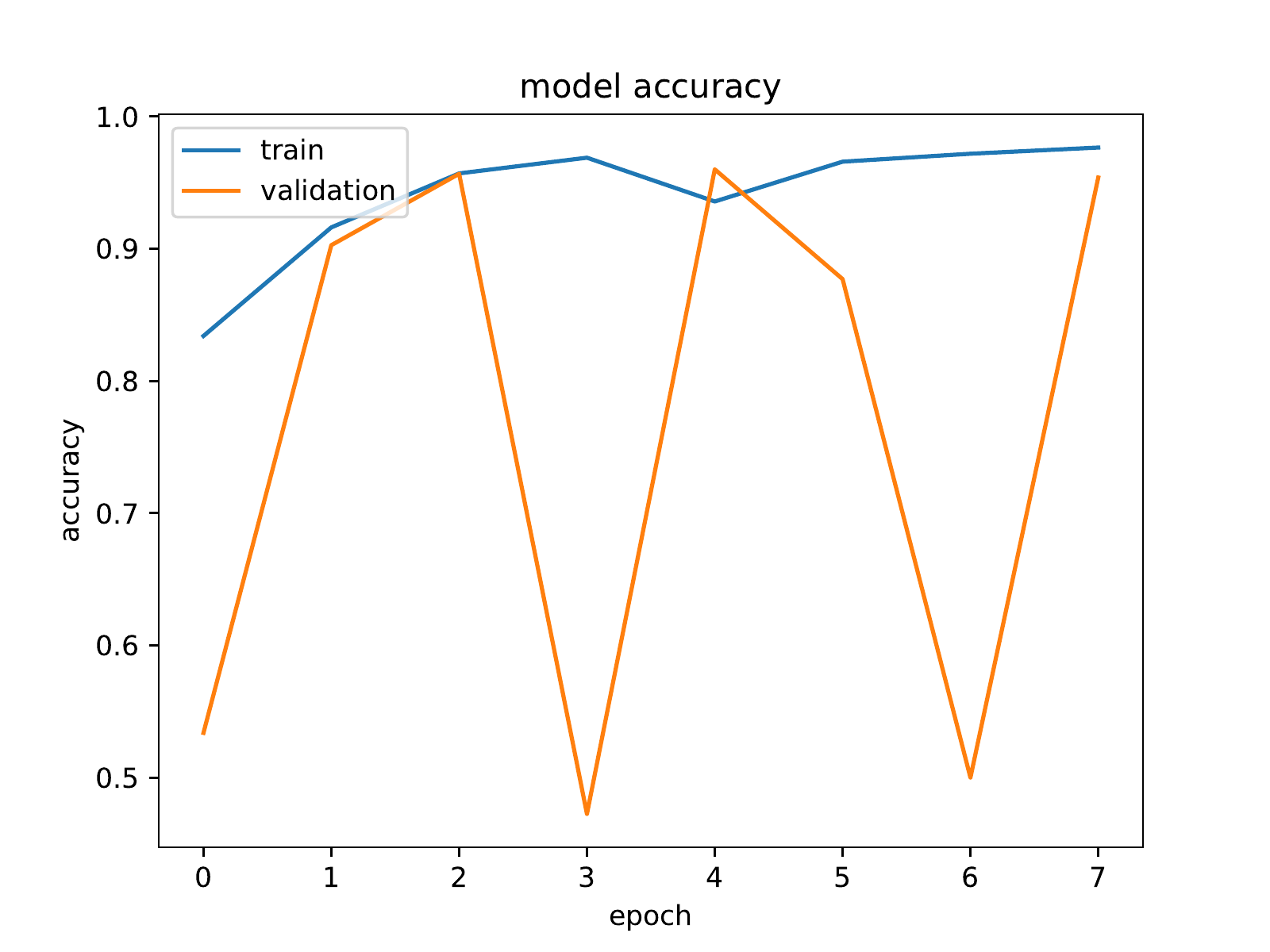}}
    \subfigure[\texttt{DENSENET \cite{marraetal:2018,HUANGETAL:2017}}]{\includegraphics[scale=0.3]{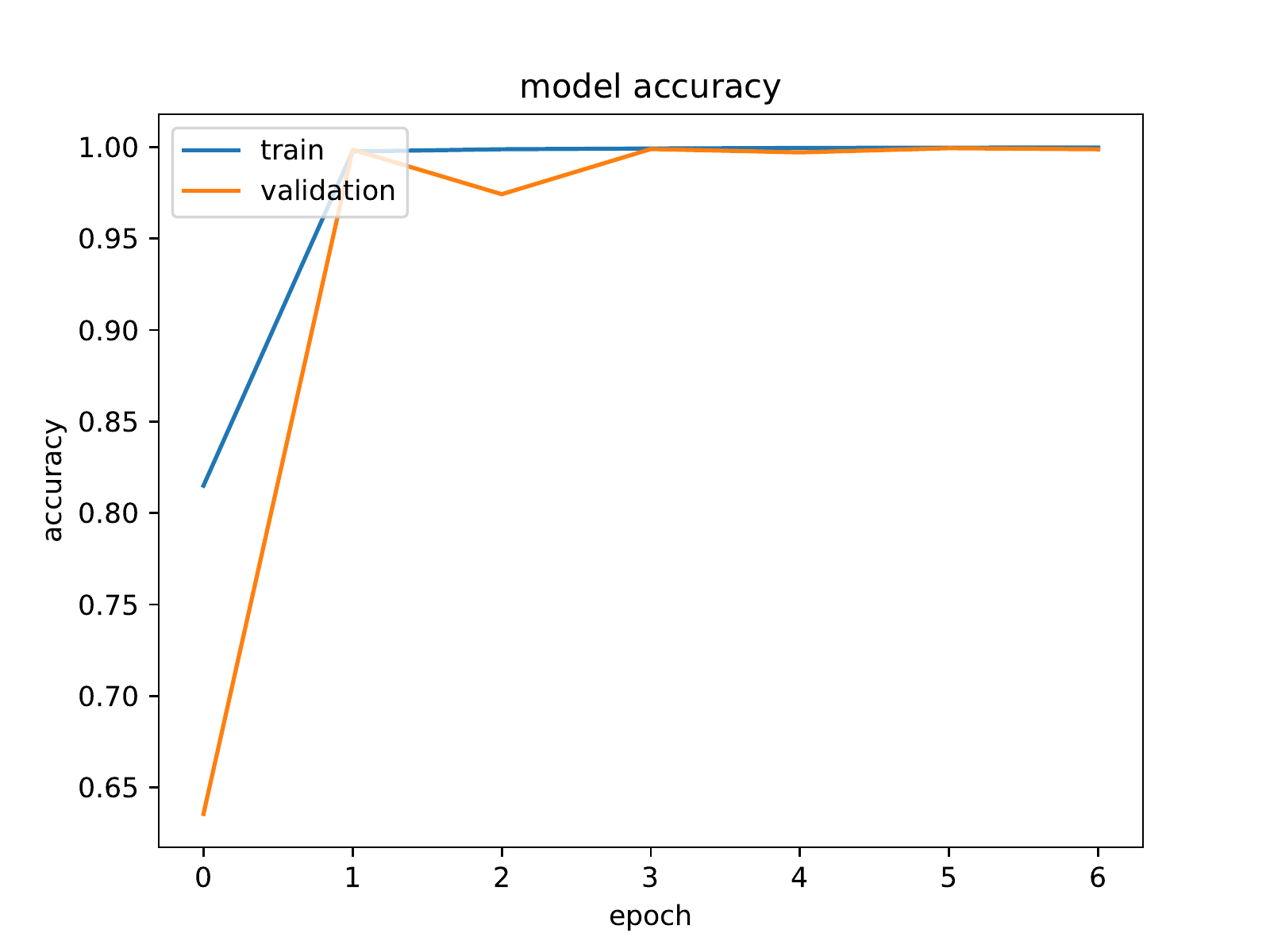}}
    \hfill
    \subfigure[\texttt{INCEPTION-V3} \cite{marraetal:2018,Szegedyetal:2016}]{\includegraphics[scale=0.3]{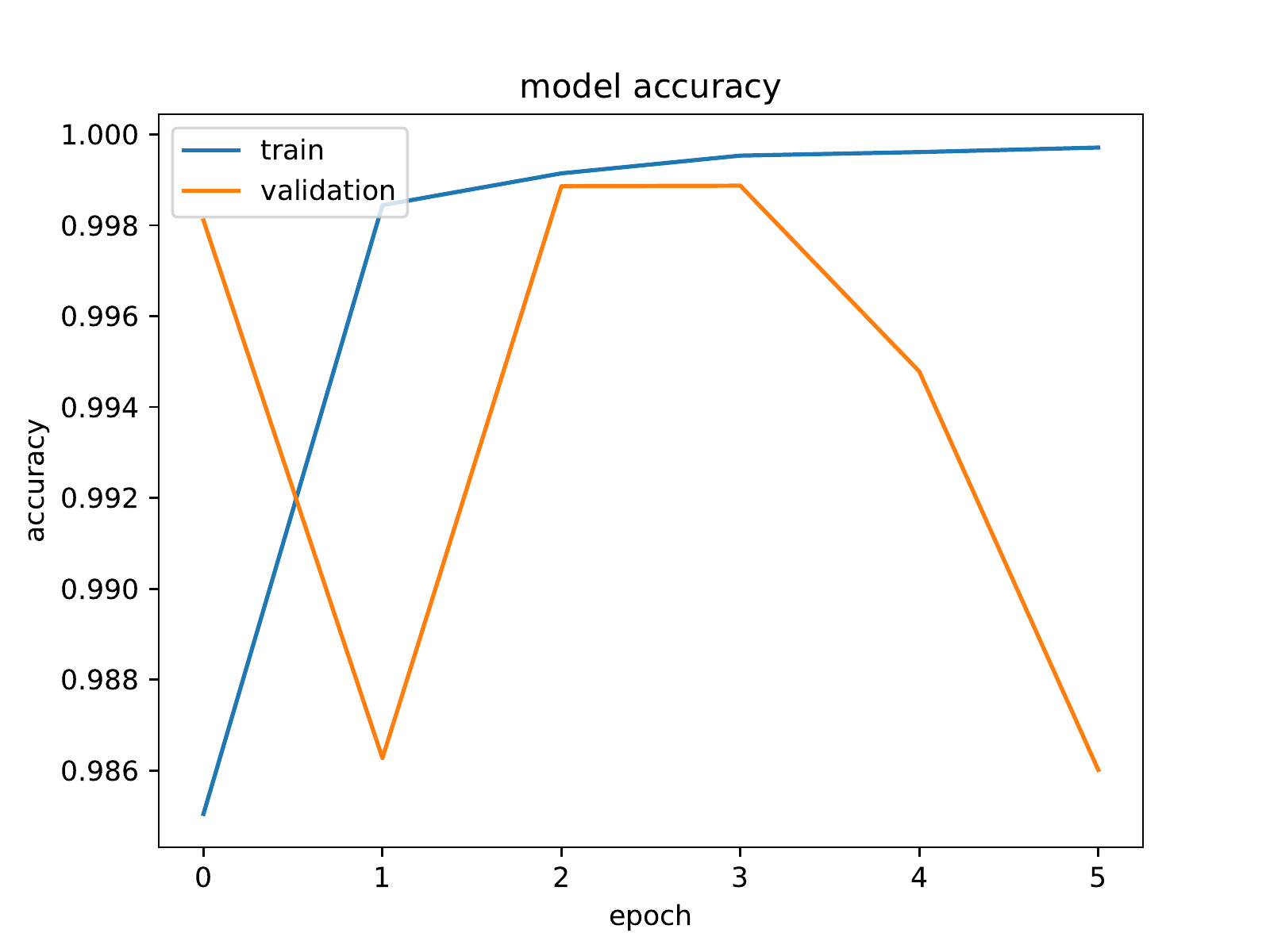}}
    \subfigure[\texttt{XCEPTION} \cite{marraetal:2018,chollet:2017}]{\includegraphics[scale=0.3]{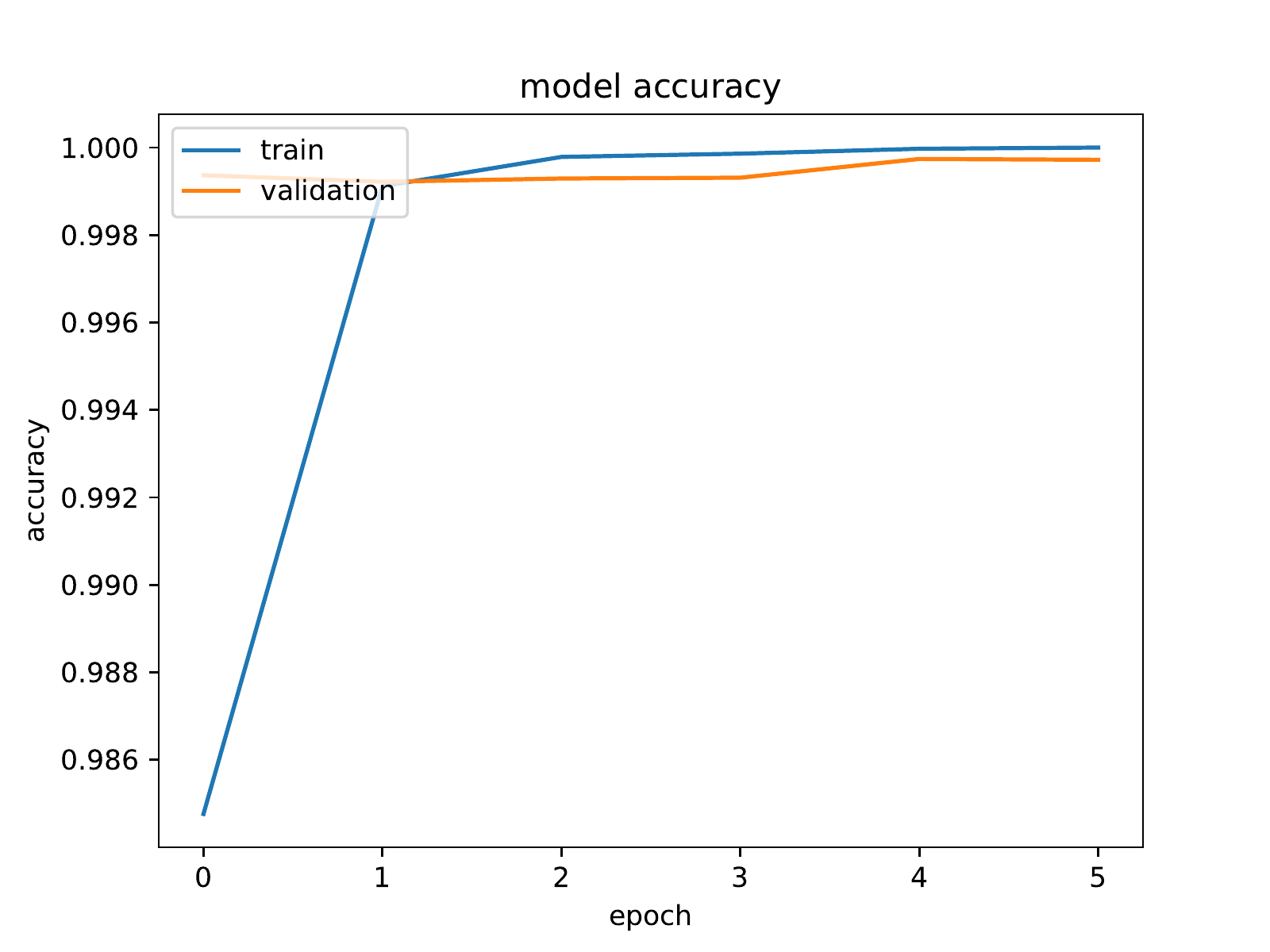}}
    \caption{Models train and validation curves when applied to digital images.}
    \label{fig:gan_trained_models}
\end{figure}

We now discuss the results of GAN image detection on printed and scanned documents. For that, we considered a set of CNNs trained on different patch sizes (\textit{i.e.,} $224 \times 224$ and $299 \times 299$) with majority voting and also $256 \times 256$ co-occurrence matrices without majority voting. We first show, in Figure \ref{fig:gan_trained_models}, the training and validation behavior of these networks considered for this experiment when trained on digital images.

The training and validation curves are shown in Figure \ref{fig:gan_trained_models} highlight the effectiveness of these networks when applied to digital images. Artifacts on image statistics and intra and inter channels pixels neighboring relations are effectively used as clues to discriminate GAN and pristine images. Indeed, all the approaches required less than 10 epochs to fulfill the early stopping criterion, finishing their training with validation accuracies higher than 95\%. 

However, as anticipated in Section \ref{deepfake-dataset}, the print and scan process eliminates most of these artifacts commonly found on digital images. In Table \ref{tab:resultsgan}, we show the classification results considering both digital and printed and scanned test images. For the digital case, all the approaches achieved an accuracy higher than 95\%, with the worst approach being \texttt{CONET} with a 96\% accuracy. The \texttt{CROSSCONET} \cite{barnietal:2020} showed better performance than \texttt{CONET} for digital images, as it also looks for artefacts in cross-band co-occurrence matrices. The best approaches in the digital scenario are \texttt{DENSENET}, \texttt{INCEPTION-V3} and \texttt{XCEPTION}, with virtually perfect results. The power of ROIs majority voting is exemplified by the confusion matrix of the \texttt{XCEPTION} CNN in Table \ref{tab:resultsxception}. It can be seen from that table that the approach misclassifies only seven $299\times299\times3$ high-energy testing patches, explaining the perfect classification after majority voting.

\begin{table}[h!]
    \centering
         \includegraphics[scale=0.89]{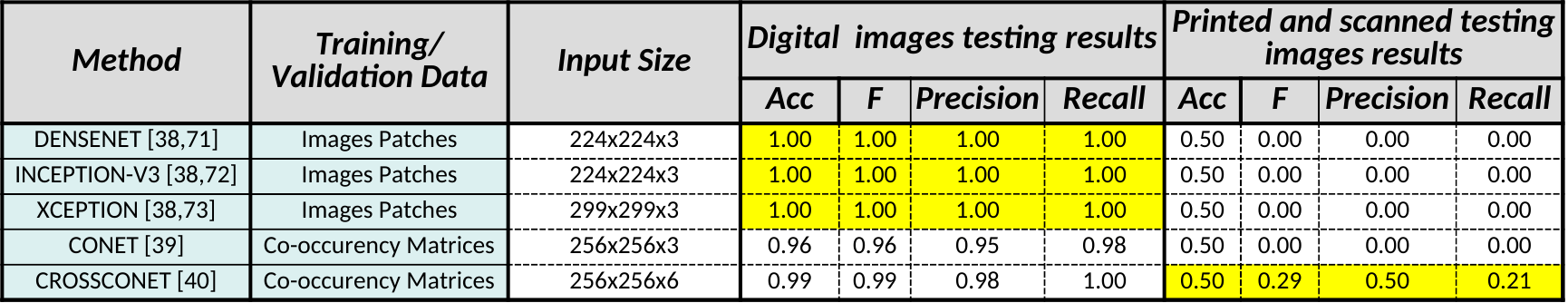} 
    \caption{Results of GAN images detection tests for digital (left) and printed and scanned images (right).}
    \label{tab:resultsgan}
\end{table}

\begin{table}[h!]
    \centering
         \includegraphics[scale=0.45]{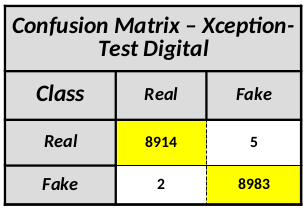} 
    \caption{Confusion matrix of \texttt{XCEPTION} $299\times299\times3$ patches classification for digital GAN-images detection.}
    \label{tab:resultsxception}
\end{table}

When faced with printed and scanned images, though, all methods fail as can be seen in the rightmost part of Table \ref{tab:resultsgan}. These results confirm that most of the artifacts used by the detectors to distinguish between GAN and real images such as warping, blur, noise, correlation, and image statistics are gone when images are printed and recaptured. 
In fact, all the approaches provide accuracy equals to 0.5 with zero precision and recall, meaning that all the images are classified as natural ones. The only minor exception is represented by \texttt{CROSSCONET} that correctly classifies 21\% of StyleGAN2 images as fake images, as it can be seen from the confusion matrix shown in Table \ref{tab:crossconet_gan_result}. 

\begin{table}[h!]
    \centering
    \includegraphics[scale=0.45]{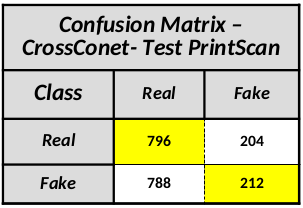}
    \caption{Confusion matrix of \texttt{CROSSCONET} for printed and scanned GAN-images detection.}
    \label{tab:crossconet_gan_result}
\end{table}

The poor results obtained when GAN-image detectors trained on digital images are applied to printed and scanned images call for new research on this topic, in order to face the fact that counterfeiters could print and scan fake images in order to avoid that they are revealed as such. We are, therefore, confident that the availability of the VIPPrint dataset will help researchers to solve this challenging task.


%% file: conclusion.tex
\section{Conclusion}
\label{conclusion}
The accessibility and constant upgrade of devices capable of generating high-quality physical documents have raised the necessity of forensics methods to attest the reliability of a printed document and possibly link illegal or criminal documents to their creator. Authentication and source linking of printed documents may also have a huge economical impact since it may help to tackle the diffusion of counterfeited products. Although several works in the scientific literature have addressed such an issue, all of them fail in two aspects: (i) they do not consider a dataset that grows with time, including more recent and professional printing devices; and (ii) they do not consider the authentication of printed artificial images.

In this paper, we show the extent of such limitations by validating existing authentication and source linking methodologies on a novel dataset specifically thought for printed documents forensics. The new dataset, the VIPPrint dataset, presents the first version of an ongoing effort to build a challenging environment for printed image forensics. To the best of our knowledge, the dataset contains the richest publicly available corpus of printed natural and artificial images, with 40,000 images addressing deepfake face-images detection, and 1,600 images focusing on source attribution in a closed set of eight printer sources. The experiments we have run showed that such dataset results in an error probability of at least 9\% for the best baseline source attribution methods. The dataset raises even more challenging problems in the case of GAN-images detection, given that StyleGAN2 images look like original ones for all the tested methods after they are printed and\ scanned.

The experiments we have run guide us to a bunch of future works. First of all, we will continue updating the dataset including new printers, more scanners, adding other GANs, and acquisition devices such as Digital Single Lens Reflex (DSLR) cameras. Second, we aim at investigating novel ways of selecting regions of interest in the digitized images and also consider other color spaces in addition to RGB. Finally, we are also headed to investigate and apply adversarial attacks in the printed domain, adding such a feature to our dataset in order to evaluate the effectiveness of printed document forensics methods.

%% file: main.bbl
\begin{thebibliography}{10}

\bibitem{eff:nodots}
{Electronic Frontier Foundation}.
\newblock List of printers which do or do not display tracking dots.
\newblock Available at
  \url{https://www.eff.org/pages/list-printers-which-do-or-do-not-display-tracking-dots}.

\bibitem{richteretal:2018}
Timo Richter, Stephan Escher, Dagmar Sch\"{o}nfeld, and Thorsten Strufe.
\newblock Forensic analysis and anonymisation of printed documents.
\newblock In {\em ACM Workshop on Information Hiding and Multimedia Security},
  page 127–138, New York, NY, USA, 2018. Association for Computing Machinery.

\bibitem{alietal:2004}
G.N. Ali, A.K. Mikkilineni, E.J. Delp, J.P. Allebach, P.-J. Chiang, and G.T.
  Chiu.
\newblock Application of principal components analysis and gaussian mixture
  models to printer identification.
\newblock In {\em International Conference on Digital Printing Technologies},
  pages 301--305, 2004.

\bibitem{mikkilinenietal:2004}
{Aravind K.} Mikkilineni, {Pei Ju} Chiang, {Gazi N.} Ali, {George T.C.} Chiu,
  {Jan P.} Allebach, and {Edward J.} Delp.
\newblock Printer identification based on texture features.
\newblock In {\em International Conference on Digital Printing Technologies},
  pages 306--311, 12 2004.

\bibitem{Mikkilinenietal:2005}
Aravind~K. Mikkilineni, Pei-Ju Chiang, Gazi~N. Ali, George T.~C. Chiu, Jan~P.
  Allebach, and Edward J.~Delp III.
\newblock {Printer identification based on graylevel co-occurrence features for
  security and forensic applications}.
\newblock In Edward J.~Delp III and Ping~W. Wong, editors, {\em Security,
  Steganography, and Watermarking of Multimedia Contents VII}, volume 5681,
  pages 430 -- 440. International Society for Optics and Photonics, SPIE, 2005.

\bibitem{Mikkilinenietal:2005a}
A.K. Mikkilineni, Osman Arslan, P.-J Chiang, R.M. Kumontoy, J.P. Allebach,
  George Chiu, and E.J. Delp.
\newblock Printer forensics using svm techniques.
\newblock In {\em International Conference on Digital Printing Technologies},
  pages 223--226, 01 2005.

\bibitem{keeandfarid:2008}
Eric Kee and Hany Farid.
\newblock Printer profiling for forensics and ballistics.
\newblock In {\em Proceedings of the 10th ACM Workshop on Multimedia and
  Security}, pages 3--10, New York, NY, USA, 2008. Association for Computing
  Machinery.

\bibitem{dengetal:2008}
W.~{Deng}, Q.~{Chen}, F.~{Yuan}, and Y.~{Yan}.
\newblock Printer identification based on distance transform.
\newblock In {\em International Conference on Intelligent Networks and
  Intelligent Systems}, pages 565--568, 2008.

\bibitem{wuetal:2009}
{Yubao Wu}, {Xiangwei Kong}, {Xin'gang You}, and {Yiping Guo}.
\newblock Printer forensics based on page document's geometric distortion.
\newblock In {\em 2009 16th IEEE International Conference on Image Processing
  (ICIP)}, pages 2909--2912, 2009.

\bibitem{jiangetal:2010}
Weina Jiang, Anthony T.~S. Ho, Helen Treharne, and Yun~Q. Shi.
\newblock A novel multi-size block benford's law scheme for printer
  identification.
\newblock In Guoping Qiu, Kin~Man Lam, Hitoshi Kiya, Xiang-Yang Xue, C.-C.~Jay
  Kuo, and Michael~S. Lew, editors, {\em Advances in Multimedia Information
  Processing - PCM 2010}, pages 643--652, Berlin, Heidelberg, 2010. Springer
  Berlin Heidelberg.

\bibitem{mikkilinenietal:2011}
Aravind~K. Mikkilineni, Nitin Khanna, and Edward~J. Delp.
\newblock {Forensic printer detection using intrinsic signatures}.
\newblock In Nasir~D. Memon, Jana Dittmann, Adnan~M. Alattar, and Edward
  J.~Delp III, editors, {\em Media Watermarking, Security, and Forensics III},
  volume 7880, pages 278 -- 288. International Society for Optics and
  Photonics, SPIE, 2011.

\bibitem{tsaiandliu:2013}
M.~{Tsai} and J.~{Liu}.
\newblock Digital forensics for printed source identification.
\newblock In {\em IEEE International Symposium on Circuits and Systems
  (ISCAS)}, pages 2347--2350, 2013.

\bibitem{elkasrawishafait:2014}
S.~{Elkasrawi} and F.~{Shafait}.
\newblock Printer identification using supervised learning for document forgery
  detection.
\newblock In {\em IAPR International Workshop on Document Analysis Systems},
  pages 146--150, 2014.

\bibitem{tsaietal:2014}
Min-Jen Tsai, Jin-Shen Yin, Imam Yuadi, and Jung Liu.
\newblock Digital forensics of printed source identification for chinese
  characters.
\newblock {\em Multimedia Tools and Applications}, 73(3):2129–2155, 12 2014.

\bibitem{haoetal:2015}
J.~{Hao}, X.~{Kong}, and S.~{Shang}.
\newblock Printer identification using page geometric distortion on text lines.
\newblock In {\em IEEE China Summit and International Conference on Signal and
  Information Processing (ChinaSIP)}, pages 856--860, 2015.

\bibitem{shangetal:2015}
Shize Shang, Xiangwei Kong, and Xingang You.
\newblock {Document forgery detection using distortion mutation of geometric
  parameters in characters}.
\newblock {\em Journal of Electronic Imaging}, 24(2):1 -- 10, 2015.

\bibitem{tsaietal:2015}
M.~{Tsai}, C.~{Hsu}, J.~{Yin}, and I.~{Yuadi}.
\newblock Japanese character based printed source identification.
\newblock In {\em IEEE International Symposium on Circuits and Systems
  (ISCAS)}, pages 2800--2803, 2015.

\bibitem{anselmoetal:2017}
A.~Ferreira, L.~Bondi, L.~Baroffio, P.~Bestagini, J.~Huang, J.~A. dos Santos,
  S.~Tubaro, and A.~Rocha.
\newblock Data-driven feature characterization techniques for laser printer
  attribution.
\newblock {\em IEEE Transactions on Information Forensics and Security},
  12(8):1860--1873, Aug 2017.

\bibitem{joshikhanna:2018}
S.~{Joshi} and N.~{Khanna}.
\newblock Single classifier-based passive system for source printer
  classification using local texture features.
\newblock {\em IEEE Transactions on Information Forensics and Security},
  13(7):1603--1614, 2018.

\bibitem{joshietal:2018}
S.~{Joshi}, M.~{Lomba}, V.~{Goyal}, and N.~{Khanna}.
\newblock Augmented data and improved noise residual-based cnn for printer
  source identification.
\newblock In {\em 2018 IEEE International Conference on Acoustics, Speech and
  Signal Processing (ICASSP)}, pages 2002--2006, 2018.

\bibitem{Jainandkhanna:2019}
Hardik Jain, G.~Gupta, S.~Joshi, and Nitin Khanna.
\newblock Passive classification of source printer using text-line-level
  geometric distortion signatures from scanned images of printed documents.
\newblock {\em Multimedia Tools and Applications}, 79:7377--7400, 2019.

\bibitem{joshikhanna:2020}
S.~{Joshi} and N.~{Khanna}.
\newblock Source printer classification using printer specific local texture
  descriptor.
\newblock {\em IEEE Transactions on Information Forensics and Security},
  15:160--171, 2020.

\bibitem{alietal:2003}
Gazi Ali, Aravind Mikkilineni, Pei-Ju Chiang, Jan Allebach, George Chiu, and
  Edward Delp.
\newblock Intrinsic and extrinsic signatures for information hiding and secure
  printing with electrophotographic devices.
\newblock In {\em International Conference on Digital Printing Technologies},
  09 2003.

\bibitem{eidetal:2008}
A.~H. {Eid}, M.~N. {Ahmed}, and E.~E. {Rippetoe}.
\newblock Ep printer jitter characterization using 2d gabor filter and spectral
  analysis.
\newblock In {\em IEEE International Conference on Image Processing}, pages
  1860--1863, 2008.

\bibitem{bulanetal:2009}
O.~{Bulan}, J.~{Mao}, and G.~{Sharma}.
\newblock Geometric distortion signatures for printer identification.
\newblock In {\em 2009 IEEE International Conference on Acoustics, Speech and
  Signal Processing}, pages 1401--1404, 2009.

\bibitem{choietal:2009}
{Jung-Ho Choi}, {Dong-Hyuck Im}, {Hae-Yeoun Lee}, {Jun-Taek Oh}, {Jin-Ho Ryu},
  and {Heung-Kyu Lee}.
\newblock Color laser printer identification by analyzing statistical features
  on discrete wavelet transform.
\newblock In {\em IEEE International Conference on Image Processing (ICIP)},
  pages 1505--1508, 2009.

\bibitem{choietal:2010}
Jung-Ho Choi, Heung-Kyu Lee, Hae-Yeoun Lee, and Young-Ho Suh.
\newblock Color laser printer forensics with noise texture analysis.
\newblock In {\em ACM Workshop on Multimedia and Security}, page 19–24, New
  York, NY, USA, 2010. Association for Computing Machinery.

\bibitem{ryuetal:2010}
S.~{Ryu}, H.~{Lee}, D.~{Im}, J.~{Choi}, and H.~{Lee}.
\newblock Electrophotographic printer identification by halftone texture
  analysis.
\newblock In {\em IEEE International Conference on Acoustics, Speech and Signal
  Processing}, pages 1846--1849, 2010.

\bibitem{tsaielat:2011}
M.~{Tsai}, J.~{Liu}, C.~{Wang}, and C.~{Chuang}.
\newblock Source color laser printer identification using discrete wavelet
  transform and feature selection algorithms.
\newblock In {\em IEEE International Symposium of Circuits and Systems
  (ISCAS)}, pages 2633--2636, 2011.

\bibitem{choietal:2013}
Jung-Ho Choi, Hae-Yeoun Lee, and Heung-Kyu Lee.
\newblock Color laser printer forensic based on noisy feature and support
  vector machine classifier.
\newblock {\em Multimedia Tools and Applications}, 67(2):363--382, Nov 2013.

\bibitem{kimandlee:2014}
D.~{Kim} and H.~{Lee}.
\newblock Color laser printer identification using photographed halftone
  images.
\newblock In {\em European Signal Processing Conference (EUSIPCO)}, pages
  795--799, 2014.

\bibitem{wuetal:2015}
H.~{Wu}, X.~{Kong}, and S.~{Shang}.
\newblock A printer forensics method using halftone dot arrangement model.
\newblock In {\em IEEE China Summit and International Conference on Signal and
  Information Processing (ChinaSIP)}, pages 861--865, 2015.

\bibitem{kimandlee:2015}
D.~{Kim} and H.~{Lee}.
\newblock Colour laser printer identification using halftone texture
  fingerprint.
\newblock {\em Electronics Letters}, 51(13):981--983, 2015.

\bibitem{anselmoetal:2015}
Anselmo Ferreira, Luiz~C. Navarro, Giuliano Pinheiro, Jefersson~A. dos Santos,
  and Anderson Rocha.
\newblock Laser printer attribution: Exploring new features and beyond.
\newblock {\em Forensic Science International}, 247(0):105 -- 125, 2015.

\bibitem{tsaietal:2017}
M.~{Tsai}, M.~{Yuadi}, Y.~{Tao}, and J.~{Yin}.
\newblock Source identification for printed documents.
\newblock In {\em International Conference on Collaboration and Internet
  Computing (CIC)}, pages 54--58, 2017.

\bibitem{bibietal:2019}
M.~{Bibi}, A.~{Hamid}, M.~{Moetesum}, and I.~{Siddiqi}.
\newblock Document forgery detection using printer source identification—a
  text-independent approach.
\newblock In {\em International Conference on Document Analysis and Recognition
  Workshops}, volume~8, pages 7--12, 2019.

\bibitem{jamesetal:2020}
Hailey James, Otkrist Gupta, and Dan Raviv.
\newblock Printing and scanning attack for image counter forensics, 2020.

\bibitem{marraetal:2018}
F.~{Marra}, D.~{Gragnaniello}, D.~{Cozzolino}, and L.~{Verdoliva}.
\newblock Detection of gan-generated fake images over social networks.
\newblock In {\em IEEE Conference on Multimedia Information Processing and
  Retrieval (MIPR)}, pages 384--389, 2018.

\bibitem{lakshmananetal:2019}
Lakshmanan Nataraj, Tajuddin~Manhar Mohammed, B.~Manjunath, Shivkumar
  Chandrasekaran, Arjuna Flenner, Md~Jawadul Bappy, and Amit Roy-Chowdhury.
\newblock Detecting gan generated fake images using co-occurrence matrices.
\newblock {\em Electronic Imaging}, 2019:532--1, 01 2019.

\bibitem{barnietal:2020}
Mauro Barni, Kassem Kallas, Ehsan Nowroozi, and Benedetta Tondi.
\newblock Cnn detection of gan-generated face images based on cross-band
  co-occurrences analysis.
\newblock In {\em IEEE Workshop on Information Forensics and Security (WIFS)},
  pages 1--6, 2020.

\bibitem{karrasetal:2019}
T.~{Karras}, S.~{Laine}, and T.~{Aila}.
\newblock A style-based generator architecture for generative adversarial
  networks.
\newblock In {\em IEEE/CVF Conference on Computer Vision and Pattern
  Recognition (CVPR)}, pages 4396--4405, 2019.

\bibitem{karrasetal:2017}
Tero Karras, Timo Aila, Samuli Laine, and Jaakko Lehtinen.
\newblock Progressive growing of gans for improved quality, stability, and
  variation, 2017.

\bibitem{choietal:2017}
Yunjey Choi, Minje Choi, Munyoung Kim, Jung-Woo Ha, Sunghun Kim, and Jaegul
  Choo.
\newblock Stargan: Unified generative adversarial networks for multi-domain
  image-to-image translation, 2017.

\bibitem{khannaetal:2006}
Nitin Khanna, Aravind~K. Mikkilineni, Anthony~F. Martone, Gazi~N. Ali, George
  T.-C. Chiu, Jan~P. Allebach, and Edward~J. Delp.
\newblock A survey of forensic characterization methods for physical devices.
\newblock {\em Digital Investigation}, 3:17 -- 28, 2006.
\newblock The Proceedings of the 6th Annual Digital Forensic Research Workshop
  (DFRWS '06).

\bibitem{khannaetal:2008}
Nitin Khanna, Aravind~K. Mikkilineni, George T.~C. Chiu, Jan~P. Allebach, and
  Edward~J. Delp.
\newblock Survey of scanner and printer forensics at purdue university.
\newblock In Sargur~N. Srihari and Katrin Franke, editors, {\em Computational
  Forensics}, pages 22--34, Berlin, Heidelberg, 2008. Springer Berlin
  Heidelberg.

\bibitem{chiangetal:2009}
Pei-Ju Chiang, Nitin Khanna, Aravind Mikkilineni, Maria Segovia, Sungjoo Suh,
  Jan Allebach, George Chiu, and Edward Delp.
\newblock Printer and scanner forensics.
\newblock {\em IEEE Signal Processing Magazine}, 26:72 -- 83, 03 2009.

\bibitem{devietal:2010}
M.~Uma Devi, C.~Raghavendra Rao, and Arun Agarwal.
\newblock A survey of image processing techniques for identification of
  printing technology in document forensic perspective.
\newblock {\em International Journal of Computer Applications}, 1(1):9--15,
  2010.

\bibitem{oliverchen:2002}
J.~Oliver and J.~Chen.
\newblock Use of signature analysis to discriminate digital printing
  technologies.
\newblock In {\em International Conference on Digital Printing Technologies},
  pages 218--222, 01 2002.

\bibitem{lampertetal:2006}
C.~H. {Lampert}, L.~{Mei}, and T.~M. {Breuel}.
\newblock Printing technique classification for document counterfeit detection.
\newblock In {\em International Conference on Computational Intelligence and
  Security}, volume~1, pages 639--644, Guangzhou, China, 2006.

\bibitem{schulzeetal:2009}
Christian Schulze, Marco Schreyer, Armin Stahl, and Thomas Breuel.
\newblock Using dct features for printing technique and copy detection.
\newblock In Gilbert Peterson and Sujeet Shenoi, editors, {\em Advances in
  Digital Forensics V}, pages 95--106, Berlin, Heidelberg, 2009. Springer
  Berlin Heidelberg.

\bibitem{schreyeretal:2009}
Marco Schreyer, C.~Schulze, A.~Stahl, and W.~Effelsberg.
\newblock Intelligent printing technique recognition and photocopy detection
  for forensic document examination.
\newblock In {\em Proceedings of Informatiktage: Fachwissenschaftlicher
  Informatik-Kongress}, pages 39--42, Bonn, Germany, 01 2009.

\bibitem{ankushetal:2010}
Ankush Roy, Biswajit Halder, and Utpal Garain.
\newblock Authentication of currency notes through printing technique
  verification.
\newblock In {\em Indian Conference on Computer Vision, Graphics and Image
  Processing}, ICVGIP '10, page 383–390, New York, NY, USA, 2010. Association
  for Computing Machinery.

\bibitem{chiangetal:2011}
P.~{Chiang}, J.~P. {Allebach}, and G.~T.~. {Chiu}.
\newblock Extrinsic signature embedding and detection in electrophotographic
  halftoned images through exposure modulation.
\newblock {\em IEEE Transactions on Information Forensics and Security},
  6(3):946--959, 2011.

\bibitem{Beusekometal:2012}
J.~V. Beusekom, F.~Shafait, and T.~Breuel.
\newblock Automatic authentication of color laser print-outs using machine
  identification codes.
\newblock {\em Pattern Analysis and Applications}, 16:663--678, 2012.

\bibitem{NOWROOZI2021102092}
Ehsan Nowroozi, Ali Dehghantanha, Reza~M. Parizi, and Kim-Kwang~Raymond Choo.
\newblock A survey of machine learning techniques in adversarial image
  forensics.
\newblock {\em Computers \& Security}, 100:102092, 2021.

\bibitem{bondietal:2017}
L.~{Bondi}, L.~{Baroffio}, D.~{Güera}, P.~{Bestagini}, E.~J. {Delp}, and
  S.~{Tubaro}.
\newblock First steps toward camera model identification with convolutional
  neural networks.
\newblock {\em IEEE Signal Processing Letters}, 24(3):259--263, 2017.

\bibitem{ferreiraetal:2018}
A.~{Ferreira}, H.~{Chen}, B.~{Li}, and J.~{Huang}.
\newblock An inception-based data-driven ensemble approach to camera model
  identification.
\newblock In {\em IEEE International Workshop on Information Forensics and
  Security (WIFS)}, pages 1--7, Dec 2018.

\bibitem{agarwaletal:2018}
S.~{Agarwal}, W.~{Fan}, and H.~{Farid}.
\newblock A diverse large-scale dataset for evaluating rebroadcast attacks.
\newblock In {\em International Conference on Acoustics, Speech and Signal
  Processing (ICASSP)}, pages 1997--2001, 2018.

\bibitem{haodongetal:2020}
Haodong Li, Bin Li, Shunquan Tan, and Jiwu Huang.
\newblock Identification of deep network generated images using disparities in
  color components.
\newblock {\em Signal Processing}, 174:107616, 2020.

\bibitem{hsuetal:2018}
C.~{Hsu}, C.~{Lee}, and Y.~{Zhuang}.
\newblock Learning to detect fake face images in the wild.
\newblock In {\em International Symposium on Computer, Consumer and Control
  (IS3C)}, pages 388--391, 2018.

\bibitem{bonettinietal:2020}
Nicolò Bonettini, Edoardo~Daniele Cannas, Sara Mandelli, Luca Bondi, Paolo
  Bestagini, and Stefano Tubaro.
\newblock Video face manipulation detection through ensemble of cnns, 2020.

\bibitem{ssi}
{Zhou Wang}, A.~C. {Bovik}, H.~R. {Sheikh}, and E.~P. {Simoncelli}.
\newblock Image quality assessment: from error visibility to structural
  similarity.
\newblock {\em IEEE Transactions on Image Processing}, 13(4):600--612, 2004.

\bibitem{Dietterich:98}
Thomas~G. Dietterich.
\newblock Approximate statistical test for comparing supervised classification
  learning algorithms.
\newblock {\em Neural Computation}, 10(7):1895--1923, 1998.

\bibitem{friedman:1937}
Milton Friedman.
\newblock The use of ranks to avoid the assumption of normality implicit in the
  analysis of variance.
\newblock {\em Journal of the American Statistical Association},
  32(200):675--701, 1937.

\bibitem{student08ttest}
William~Sealy Gosset.
\newblock The probable error of a mean.
\newblock {\em Biometrika}, 6(1):1--25, March 1908.
\newblock Originally published under the pseudonym ``Student''.

\bibitem{jainvailaya:1996}
Anil~K. Jain and Aditya Vailaya.
\newblock Image retrieval using color and shape.
\newblock {\em Pattern Recognition}, 29(8):1233 -- 1244, 1996.

\bibitem{dalatriggs:2005}
N.~{Dalal} and B.~{Triggs}.
\newblock Histograms of oriented gradients for human detection.
\newblock In {\em IEEE Computer Society Conference on Computer Vision and
  Pattern Recognition (CVPR)}, volume~1, pages 886--893, 2005.

\bibitem{Ojalaetal:96}
Timo Ojala, Matti Pietikäinen, and David Harwood.
\newblock A comparative study of texture measures with classification based on
  featured distributions.
\newblock {\em Pattern Recognition}, 29(1):51--59, 1996.

\bibitem{Simonyan15}
Karen Simonyan and Andrew Zisserman.
\newblock Very deep convolutional networks for large-scale image recognition.
\newblock In {\em International Conference on Learning Representations}, 2015.

\bibitem{heetal:2016}
K.~{He}, X.~{Zhang}, S.~{Ren}, and J.~{Sun}.
\newblock Deep residual learning for image recognition.
\newblock In {\em IEEE Conference on Computer Vision and Pattern Recognition
  (CVPR)}, pages 770--778, 2016.

\bibitem{HUANGETAL:2017}
G.~{Huang}, Z.~{Liu}, L.~{Van Der Maaten}, and K.~Q. {Weinberger}.
\newblock Densely connected convolutional networks.
\newblock In {\em 2017 IEEE Conference on Computer Vision and Pattern
  Recognition (CVPR)}, pages 2261--2269, 2017.

\bibitem{Szegedyetal:2016}
C.~{Szegedy}, V.~{Vanhoucke}, S.~{Ioffe}, J.~{Shlens}, and Z.~{Wojna}.
\newblock Rethinking the inception architecture for computer vision.
\newblock In {\em 2016 IEEE Conference on Computer Vision and Pattern
  Recognition (CVPR)}, pages 2818--2826, 2016.

\bibitem{chollet:2017}
F.~{Chollet}.
\newblock Xception: Deep learning with depthwise separable convolutions.
\newblock In {\em IEEE Conference on Computer Vision and Pattern Recognition
  (CVPR)}, pages 1800--1807, 2017.

\bibitem{ferreirasourcecode:2014}
Anselmo Ferreira.
\newblock Printer forensics source code.
\newblock Available at
  \url{https://github.com/anselmoferreira/printer_forensics_source_code}, 2014.

\bibitem{Wardi1988}
Y.~Wardi.
\newblock A stochastic steepest-descent algorithm.
\newblock {\em Journal of Optimization Theory and Applications},
  59(2):307--323, Nov 1988.

\end{thebibliography}
